\documentclass[sigconf,natbib=true]{acmart}
\AtBeginDocument{%
  \providecommand\BibTeX{{%
    \normalfont B\kern-0.5em{\scshape i\kern-0.25em b}\kern-0.8em\TeX}}}

\setcopyright{acmcopyright}
\copyrightyear{2024}
\acmYear{2024}
\acmBooktitle{ }

\usepackage{multirow}
\usepackage{tabularx}
\usepackage{adjustbox}
\usepackage[normalem]{ulem}
\useunder{\uline}{\ul}{}
\usepackage{makecell}
\usepackage{color}
\usepackage{subfig}
\usepackage{graphicx}
\usepackage{colortbl}

\begin{document}

%%
%% The "title" command has an optional parameter,
%% allowing the author to define a "short title" to be used in page headers.
\title{Untargeted Adversarial Attack on Knowledge Graph Embeddings}

%%
%% The "author" command and its associated commands are used to define
%% the authors and their affiliations.
%% Of note is the shared affiliation of the first two authors, and the
%% "authornote" and "authornotemark" commands
%% used to denote shared contribution to the research.
% \author{Ben Trovato}
%\authornote{Both authors contributed equally to this research.}
%\email{trovato@corporation.com}
%\orcid{1234-5678-9012}
%\author{G.K.M. Tobin}
%\authornotemark[1]
%\email{webmaster@marysville-ohio.com}
%\affiliation{%
%  \institution{Institute for Clarity in Documentation}
%  \streetaddress{P.O. Box 1212}
%  \city{Dublin}
%  \state{Ohio}
%  \country{USA}
%  \postcode{43017-6221}
%}

\author{Tianzhe Zhao}
\authornote{Corresponding author.}
\affiliation{%
  \institution{School of Computer Science and Technology, Xi'an Jiaotong University}
  \city{Xi'an}
  \country{China}
}
\email{ztz8758@foxmail.com}

\author{Jiaoyan Chen}
\affiliation{%
  \institution{Department of Computer Science, The University of Manchester}
  \city{Manchester}
  \country{United Kindom}
}
\email{jiaoyan.chen@manchester.ac.uk}

\author{Yanchi Ru}
\affiliation{%
  \institution{School of Computer Science and Technology, Xi'an Jiaotong University}
  \city{Xi'an}
  \country{China}
}
\email{2196123580@stu.xjtu.edu.cn}

\author{Qika Lin}
\affiliation{%
  \institution{National University of Singapore}
  % \city{Singapore}
  \country{Singapore}
}
\email{linqika@nus.edu.sg}

\author{Yuxia Geng}
\affiliation{%
  \institution{School of Computer Science, Hangzhou Dianzi University}
  \city{Hangzhou}
  \country{China}
}
\email{yuxia.geng@hdu.edu.cn}

\author{Jun Liu}
\affiliation{%
  \institution{National Engineering Lab for Big Data Analytics, Xi’an Jiaotong University}
  \city{Xi'an}
  \country{China}
}
\email{liukeen@xjtu.edu.cn}

%%
%% By default, the full list of authors will be used in the page
%% headers. Often, this list is too long, and will overlap
%% other information printed in the page headers. This command allows
%% the author to define a more concise list
%% of authors' names for this purpose.
\renewcommand{\shortauthors}{Zhao, et al.}

%%
%% The abstract is a short summary of the work to be presented in the
%% article.
\begin{abstract}
  Knowledge graph embedding (KGE) methods have achieved great success in handling various knowledge graph (KG) downstream tasks. However, KGE methods may learn biased representations on low-quality KGs that are prevalent in the real world. Some recent studies propose adversarial attacks to investigate the vulnerabilities of KGE methods, but their attackers are target-oriented with the KGE method and the target triples to predict are given in advance, which lacks practicability. In this work, we explore untargeted attacks with the aim of reducing the global performances of KGE methods over a set of unknown test triples and conducting systematic analyses on KGE robustness.  
  Considering logic rules can effectively summarize the global structure of a KG, we develop rule-based attack strategies to enhance the attack efficiency. 
  In particular, we consider adversarial deletion which learns rules, applying the rules to score triple importance and delete important triples, and adversarial addition which corrupts the learned rules and applies them for negative triples as perturbations. 
  Extensive experiments on two datasets over three representative classes of KGE methods demonstrate the effectiveness of our proposed untargeted attacks in diminishing the link prediction results. 
  And we also find that different KGE methods exhibit different robustness to untargeted attacks. For example, the robustness of methods engaged with graph neural networks and logic rules depends on the density of the graph. But rule-based methods like NCRL are easily affected by adversarial addition attacks to capture negative rules. 
\end{abstract}

\ccsdesc[500]{Computing methodologies~Knowledge representation and reasoning}
%%
%% Keywords. The author(s) should pick words that accurately describe
%% the work being presented. Separate the keywords with commas.
\keywords{Knowledge Graph Embedding, Knowledge Graph Completion, Adversarial Attack, Robustness, Rule Learning}

%% A "teaser" image appears between the author and affiliation
%% information and the body of the document, and typically spans the
%% page.
%%\begin{teaserfigure}
%%  \includegraphics[width=\textwidth]{sampleteaser}
%%  \caption{Seattle Mariners at Spring Training, 2010.}
%%  \Description{Enjoying the baseball game from the third-base
%%  seats. Ichiro Suzuki preparing to bat.}
%%  \label{fig:teaser}
%%\end{teaserfigure}

% \received{20 February 2007}
% \received[revised]{12 March 2009}
% \received[accepted]{5 June 2009}

%%
%% This command processes the author and affiliation and title
%% information and builds the first part of the formatted document.
\maketitle

\section{Introduction}
Knowledge Graphs (KGs) have gained wide attention for representing graph-structured data as collections of relational facts in the form of triples, i.e., \textit{(head, relation, tail)}. They play a critical role in various domains such as information retrieval~\cite{10.1145/2600428.2609628,10.1145/3209978.3210187}, recommendation systems \cite{10.1145/3331184.3331203,10.1145/3397271.3401051,DBLP:journals/tkde/ZhuLLSLCWN23} and question answering \cite{10.1145/3331184.3331252,saxena-etal-2020-improving}. With the rapid advancement of representation learning, many knowledge graph embedding (KGE) methods (such as TransE~\cite{bordes_2013}, ComplEx \cite{pmlr-v48-trouillon16}, ConvE \cite{dettmers2018convolutional} and CompGCN \cite{Vashishth2020Composition-based}) have been proposed to project entities and relations into a continuous vector space for facilitating efficient incorporation of KGs in downstream applications especially with machine learning and data mining.
\begin{figure*}[h]
  \centering
  \includegraphics[width=0.8\linewidth]{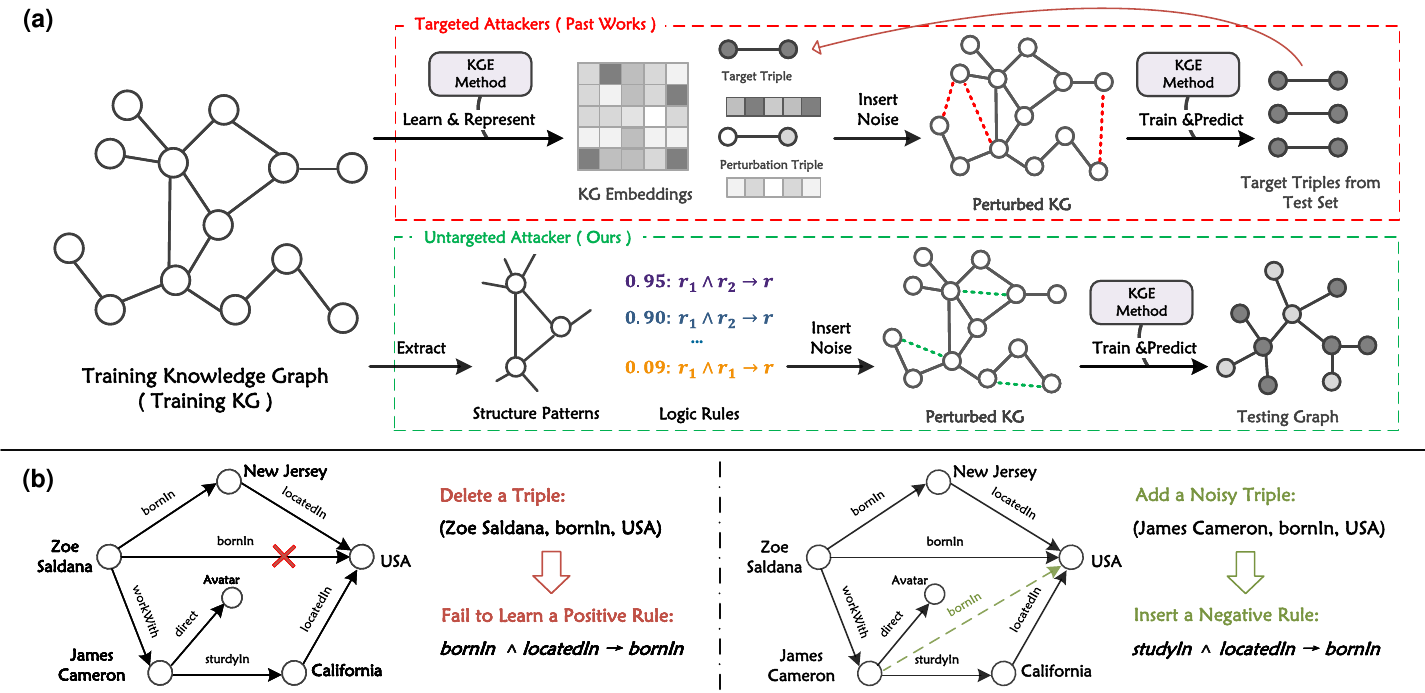}
  \caption{(a) Illustration of targeted and untargeted adversarial attacks on KGE. (b) Illustration of KG perturbation examples that can lead to positive rules missing and negative rules learned, respectively.}
  \label{fig:intro}
  % \Description{}
  % \setlength{\abovecaptionskip}{-0.5cm}
  % \setlength{\belowcaptionskip}{-0.4cm}  
  \vspace{-0.2cm}
\end{figure*} 

Despite the increasing attention on KGE, several researches~\cite{ijcai2019p674,pezeshkpour-etal-2019-investigating} have also witnessed that KGE methods are often fragile to KGs with low quality which are common in the real world. On the one hand, KGs may be polluted on purpose. Many deployed KGs like cybersecurity KGs \cite{kiesling2019sepses,wang2021social,sikos2023cybersecurity} are under frequent malicious attacks. Some open KGs collect data from public resources \cite{10.1145/3543507.3583203} and can be edited by humans such as Wikidata \cite{vrandevcic2014wikidata}. It is possible that attackers make malicious editions on these KGs.
On the other hand, data perturbations may occur during automatic KG construction \cite{yu2020relationship,10.1145/3404835.3463113,zheng-etal-2021-prgc}. It may extract incorrect triples from the raw data sources and miss correct ones due to data incompleteness. With such disruption and quality issues, KGE methods are much more likely to learn biased embeddings leading to serious negative impacts on real-world applications. 
For example, in a KG-based recommendation system, once a wrong category is added to an item, this item may be recommended to users who have no interest on it. Also, as for the entity query in information retrieval, deleting useful triples may lead to incomplete query results, like removing the triple \textit{(James Cameron, direct, Titanic)} might result in missing the correct answer \textit{Titanic} to the query \textit{Which movies directed by James Cameron have won the Academy Award?}
Therefore, investigating KGE methods when the KG suffers from attacks or quality issues, and understanding their robustness are relatively significant.

Some recent studies employ adversarial attacks on KGs by adding perturbation triples or deleting triples to make KGE methods fail in predicting the preset target facts\cite{ijcai2019p674,DBLP:conf/emnlp/BhardwajKCO21,10.1145/3543507.3583203}. 
These studies have full access to certain target triples and develop attacks towards reducing the testing performance on these certain triples. Such attacks are known as target-oriented attacks. Moreover, the existing attackers acquire full knowledge about the KGE method to attack, including the KG embeddings it learns, as illustrated in Figure \ref{fig:intro} (a).
To be more specific, Bhardwaj et al. \cite{DBLP:conf/emnlp/BhardwajKCO21} add noisy triples with high instance-level similarities to the explicit targets and delete triples with low similarities. Similarly, targeted attackers can also poison a KG according to the scores of target and perturbed triples~\cite{ijcai2019p674}. Besides, several works enhance the efficiency of such target-oriented attacks by implementing an influence function \cite{pezeshkpour-etal-2019-investigating} and relation inference patterns \cite{DBLP:conf/acl/BhardwajKCO20} which are based on the given KG embeddings.

Although these attacks can successfully reduce the performance of some KGE methods over the selected targets, they have two main limitations. 
Firstly, target-oriented attacks lack practicability in real-world scenarios. For instance, in a KG-based online recommendation system, it is not realistic to know the item-user pairs for recommendation when the KG embeddings are learned.
Secondly, attacks specific to a KGE method cannot help explain the general regularities of the KG. 
Aside from performance decrease over the evaluation task, we expect to figure out how attacks really impact the KG and its embeddings, thus developing more robust KGE methods. Attacks relying on a certain KGE method like TransE are more likely to reflect the features of the KGE method itself rather than the properties of the KG. Such attacks may also fail on other KGE methods. Besides, these methods develop attacks by correlation analysis of the embeddings, which cannot provide explanations.

To overcome these limitations, we propose untargeted KGE attacks which cannot access either the testing triples or the KGE method, and aim at reducing the global performance on all the testing triples as shown in Figure \ref{fig:intro} (a). With such attacks, we further investigate the vulnerability of KGE methods systematically.
Unlike targeted attacks that have target triples as direct references, to develop untargeted attack, we need to trace the influence of each triple on the KG structure. Motivated by \cite{10.1145/3308558.3313612,10.1145/3534678.3539338,DBLP:conf/ijcai/WangWG15,zhang2022knowledge,10.1145/3477495.3531996}, where logic rules have demonstrated the great power in reasoning over KGs with explanations, we find that 
logic rules can well describe the knowledge of a KG and a KG's structure is highly associated with its rules. When the KG structure changes as some critical triples are added or deleted, the rules will correspondingly change, and vice versa. Take Figure \ref{fig:intro} (b) for instance, removing the triple \textit{(Zoe Saldana, bornIn, USA)} disrupts a closed path and may make it unable to derive the positive rule ${bornIn(X,Z)}\wedge{locatedIn(Z,Y)}\rightarrow{bornIn(X,Y)}$. Also, adding a noisy triple \textit{(James Cameron, bornIn, USA)} may result in the occurrence of the negative rule ${studyIn(X,Z)}\wedge{locatedIn(Z,Y)}\rightarrow{bornIn(X,Y)}$.
With such observation, we consider the importance of each triple via its contribution to support some logic rules and propose to develop untargeted attacks using KG's logic rules.

To this end, we first apply a rule learning method to extract rules from the KG, which serve as the prior knowledge that reflects its structure. We then develop untargeted attacks to poison the general knowledge of the KG to the greatest extent in both deletion and addition settings. For deletion, we remove triples which support the grounding of high-confident logic rules. With the absence of such triples, many trustworthy logic rules will be missing so that the global feature of the KG can be fragmented. In the addition attack, added perturbations are expected to mislead KGE methods capturing unreliable logic rules rather than reliable ones. Therefore, we disrupt the extracted logic rules with low confidence values to get a series of negative rules, and then determine the perturbation triples via inferences of the negative rules over the KG. After employing both deletion and addition attacks, we verify the attack efficiency over seven dominant KGE methods covering three typical categories: fact-based, graph neural network (GNN)-based and rule-based methods. Furthermore, we conduct comprehensive comparisons and analyses for the robustness of these KGE methods. 
 
% The main contributions of this work can be summarised as follows:
The main contributions of this work can be summarized as:

\noindent $\bullet$ We propose a new task of untargeted adversarial attack on KGE.
It is more practical than previous targeted adversarial attack with fewer assumptions and enables the analysis of KGE robustness.
%To the best of our knowledge, it is the first work to investigate the global performances of KGEs via adversarial attack, which is more practical to real-world scenarios.

\noindent $\bullet$ We develop strategies of adversarial deletion and addition for effective untargeted attacks, utilizing the KG's logic rules. 

\noindent $\bullet$ We evaluate our attacks on two KG benchmarks (FB15k-237 and WN18RR) among seven representative KGE methods. The comparison results with recent state-of-the-art targeted attack approaches demonstrate the effectiveness of our proposed methods.
% For example, our deletion attack achieves the greatest reduction on Hits@10 in FB15k-237 for all tested KGE methods.

\noindent $\bullet$ We conduct detailed analyses on the robustness of KGE methods, where most KGE methods exhibit stronger robustness to the adversarial addition than to the adversarial deletion except for TransE. The robustness of those methods that incorporate GNNs and rules heavily relies on the density of the graph structure, and these methods show robust resilience to the addition attack. 

\begin{figure*}[h]
  \centering
  \includegraphics[width=0.85\linewidth]{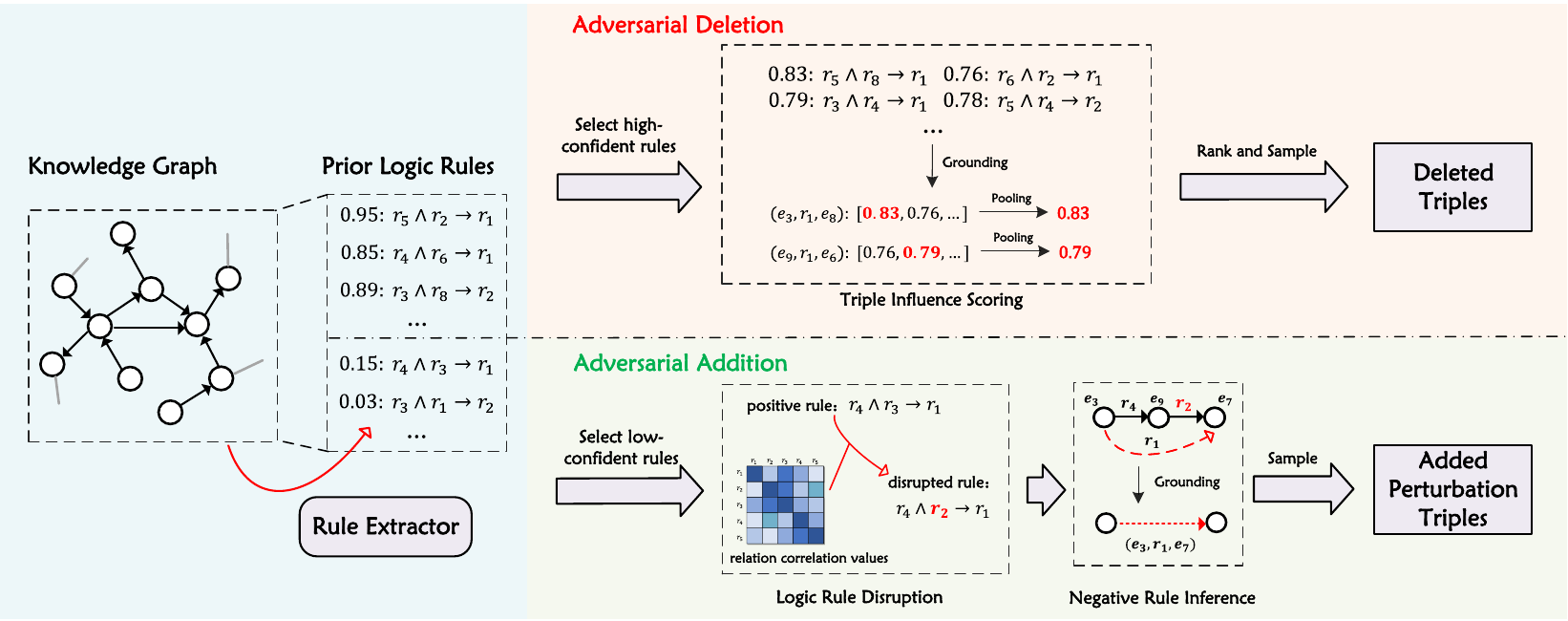}
  \caption{The overall framework of our proposed untargeted adversarial attacks.}
  \label{fig:framework}
\end{figure*} 

\section{Methodology}
In this section, we first introduce some preliminaries about background concepts and the problem definition, then present the details of our attack strategies of both deletion and addition settings. The overall framework of our proposed attacks is illustrated in Figure \ref{fig:framework}.

\subsection{Preliminary}
\subsubsection{Knowledge Graph (KG)}
In this study, we aim at KGs that are composed of relational facts in form of
RDF\footnote{Resource Description Framework. \url{https://www.w3.org/RDF/}}
triples. A KG is denoted as 
$\mathcal{G}=\{\mathcal{E},\mathcal{R},\mathcal{T}\}$, where $\mathcal{E}$, $\mathcal{R}$ and $\mathcal{T}$ represent sets of entities, relations and triples, respectively. Each triple in $\mathcal{T}$ is denoted as $(e_h,r,e_t)$ with $r\in\mathcal{R}$, $e_h\in\mathcal{E}$ and $e_t\in\mathcal{E}$, representing a relation $r$ links a head entity $e_h$ to a tail entity $e_t$. 

\subsubsection{Logic Rule}
Logic rules can be learned from a KG and applied for inferring new triples.
We study logic rules that are categorized as first-order Horn rules \cite{wu_et_al:TGDK.1.1.7,DBLP:conf/acl/LinL0XC23}, formally expressed as: 
%Specifically, relations and entities in KGs act as predicates and variables to form atoms within these rules. 
%Formally, a first-order Horn rule is expressed as:
\begin{equation}
    \alpha: r_1(X,Z_1)\wedge{r_2(Z_1,Z_2)}\wedge...\wedge{r_n(Z_{n-1},Y)}\rightarrow{r_{hd}(X,Y)},
\end{equation}
\noindent where $X$, $Z_1$, $Z_2$, ..., $Z_{n-1}$ and $Y$ denote variables that can be assigned with the KG entities, $r_1$, $r_2$, ..., $r_n$ and $r_{hd}$ denote predicates that can be assigned with the KG relations. The conjunction of the atoms $r_1(X,Z_1)$, $r_2(Z_1, Z_2)$, ...,$r_n(Z_{n-1},Y)$ composes the rule's body, while the atom $r_{hd}(X,Y)$ is the rule's head. $\alpha\in[0,1]$ denotes the rule's confidence value. For convenience, we simplify the denotation of the rule in Equation (1) as $(r_{hd}, \mathbf{r_b})$  and accordingly represent the confidence value as $\alpha(r_{hd}, \mathbf{r_b})$, with $\mathbf{r_b}=[r_1,r_2,...,r_n]$ denoting the predicate sequence of the rule's body. 
%The rule can be also represented in an intuitive manner: $r_1 \wedge r_2, ..., \wedge r_n \rightarrow r_{hd}$.

The procedure of applying a rule to a KG is called \textit{grounding}.
In this procedure, each rule variable is assigned to a specific entity, the predicates are replaced by relations, the atoms become facts such as \textit{bornIn(Zoe Saldana, USA)} that is equivalent to the triple \textit{(Zoe Saldana, bornIn, USA)}, and combinations of triples that satisfy the rule are searched. We call the grounded triples from the body atoms of a rule as its \textit{body triples}, and the grounded triples from the head atoms of a rule as its \textit{head triples}.

\subsubsection{Knowledge Graph Embedding (KGE)} 
KGE methods intend to represent entities and relations in a continuous vector space with their semantics,
% , such as relationships in the graph structure,
concerned with supporting downstream tasks that use vectors as input. 
They usually use a scoring function $\varphi$ to model the likelihood of a triple $(e_h,r,e_t)\in\mathcal{T}$ based on some preset assumption. 
For example, the classic KGE method TransE assumes the relation $r$ translates the head entity $h$ to the tail entity $t$ in the vector space, and thus defines the score function as $\varphi(e_h,r,e_t)=-\vert\vert\mathbf{e_h}+\mathbf{r}-\mathbf{e_t}\vert\vert$, where $\mathbf{e_h}$, $\mathbf{r}$, $\mathbf{e_t}$ denotes the vector presentations of $h$, $r$ and $t$, respectively, and $\vert\vert \cdot \vert\vert$ denotes the calculation of 2-norm.

\subsubsection{Problem Definition} 
KGE is widely applied to and evaluated by the task of Link Prediction, a.k.a., KG completion (KGC), which is to predict the head entity given the relation and tail entity, denoted as $(?,r,e_t)$, or to predict the tail entity given the relation and head entity, denoted as $(e_h,r,?)$. 
In particular, a KG $\mathcal{G}=\{\mathcal{E},\mathcal{R},\mathcal{T}\}$ is given for training a KGE method, and a set of ground truth triples $\mathcal{T}_{te}$ with $\mathcal{T} \cap \mathcal{T}_{te} = \emptyset$ is given for testing. A KGE method is regarded to have better performance if the triples in $\mathcal{T}_{te}$ have a higher overall score using its learned embeddings. 
In evaluation, ranking-based metrics such as Mean Reciprocal Rank (MRR) and Hits@K, which have higher scores if the entities that lead to the ground truth triples are at more forward positions when ranked together with other candidate entities, are often used to measure the performance.

The untargeted attack in our study aims to change the KG for training $\mathcal{G}$ by adding a fixed number of triples to $\mathcal{T}$ or by deleting a fixed number of triples in $\mathcal{T}$, such that the performance of a specific set of KGE methods on $\mathcal{T}_{te}$ is reduced to the maximum extent. The KG after attack is denoted as $\widehat{\mathcal{G}}=\{\mathcal{E},\mathcal{R},\widehat{\mathcal{T}}\}$, and the number of triples to add or delete is denoted as $\delta$.
``Untargeted'' means that none of the testing triples $\mathcal{T}_{te}$ is accessible to the attack procedure which finds out the triples for addition or deletion, and the attack aims to reduce the global performance on all the testing triples.
On the one hand, this matches many real-world situations where the application scenarios of the trained embeddings are unknown to the attackers. On the other hand, studying such attacks allows us to study and compare the robustness of KGE methods when the training KG is noisy or polluted.

\subsection{Rule Extraction}
To completely summarize the knowledge and represent the global structure for a KG, we expect to extract a series of logical rules with the predicates in the rule heads covering as many relations in the KG as possible. To this end, we employ a recent KG rule learning method named NCRL \cite{cheng2023neural}.
NCRL focuses on the composition structure of a rule body to capture its hierarchical nature and achieves state-of-the-art performances in KGC. Briefly, for one head entity and one tail entity in a potential triple, NCRL samples paths connecting them by a random walk algorithm, extracts a sequence of relations from each sampled path as the body, and selects correct relations between the three entities as the rule head. 

In practice, NCRL merges the composition in the sampled path and reduce the long rule body in an iterative way. Each recurrent stage includes three main steps. Take the first for example: 

$\bullet$ \textbf{Sliding Window Segmentation}.
In this step, the relation sequence is segmented into a series of windows via a sliding window mechanism, where each window represents a potential composition between adjacent relations. For example, setting the fixed window size $s=2$, the $i$-th window $w_i$ includes two adjacent relations as $w_i=[r_i,r_{i+1}]$, which may be composed into one relation later. Given the fixed window size $w$, the extracted relation sequence $\mathbf{r_b}=[r_1,r_2,...,r_n]$ can be transformed into $[w_1,...,w_{n+1-s}]$.

%The extracted relation sequence $\mathbf{r_b}=[r_1,r_2,...,r_n]$ is input for a sliding window splitting to potential compositions as $[w_1,...,w_{n+1-s}]$, where $s$ is a fixed window size.  
%NCRL encodes these potential compositions by a RNN as: 
$\bullet$ \textbf{Composition Selection}.
With all possible compositions in the relation sequence, NCRL utilizes Recurrent Neural Networks (RNNs) to encode each window (i.e., potential composition),
% get the representations of them. 
and then selects the sliding window to determine which composition is more likely to occur:
\begin{equation}
    % \vspace{+0.4cm}
    \mathbf{w_i}=SELECT[\mathbf{w_1},...,\mathbf{w_{n+1-s}}],
    % \vspace{-0.4cm}
\end{equation}
where $\mathbf{w_1},...,\mathbf{w_{n+1-s}}$ denotes the representation of $w_1,...w_{n+1-s}$ encoded by RNNs, respectively. $\mathbf{w_i}$ is utilized as the representation of the selected window, and $SELECT$ denotes the selection module which mainly consists of a fully connected neural network.

$\bullet$ \textbf{Attention-based Induction}.
Having determined which relations are to be composed together, a cross attention unit is applied to compute the attention scores of the selected window to all existing relations in the KG. Further, $\mathbf{w_i}$ is transformed into a single relation with the representation $\hat{w_i}$, which is the weighted assumption of the existing relations' embeddings.

By implementing recurrent stages recursively, NCRL predicts the rule heads based on the attention values of the final composition of the relation sequence and all relations. The attention scores are viewed as confidences of specific rules.
It's noticeable that for each extracted relation sequence (i.e., rule body), NCRL can select all existing relations as the rule head and use the attention score obtained during the learning process as the confidence of the rules. In this way, each relation can be learned as the rule head. Therefore, NCRL finally selects top-$k$ rules with the highest scores for each rule head as learned rules. 
Employing NCRL on the training KG $\mathcal{G}$, we obtain a collection of logic rules, denoted as $\Psi$ whose heads cover all relations in $\mathcal{G}$. 

\subsection{Adversarial Deletion} 
For the deletion attack, we aim at only deleting certain triples which have the most contribution to the KG structure according to the extracted logic rules. In particular, we focus on those logic rules assigned with high confidence values, since they are more likely to represent accurate and reliable patterns within the KG. Practically, for each $r\in\mathcal{R}$ serving as the head predicate in rules in $\Psi$, we select top-$m$ rules with the highest confidence values to construct the candidate rule set $\Psi_m$ for adversarial deletion. We develop an \textit{influence} scoring system that quantifies the semantic significance of each triple and removes triples assigned with high influence scores to disrupt the integration of core rules to the greatest extent and further compromise the global semantic pattern of the KG. 

Each rule $(r_{hd},\mathbf{r_{b}})$ in $\Psi_m$ is grounded to the KG, leading to its body triples and head triples.
For a triple $(e_h,r,e_t)$, if it is among the head triples, this triple's contribution score to this rule, denoted as $f((e_h,r,e_t), (r_{hd},\mathbf{r_{b}}))$, is assigned with the rule's confidence. Otherwise, $f((e_h,r,e_t), (r_{hd},\mathbf{r_{b}})) = 0$.

As for the influence values of $(e_h,r, e_t)$ to the whole KG, we define it as the aggregation of confidence values of all the highly confident rules in $\Psi_h$ as: 
\begin{equation}
    F(e_h,r,e_t) = Pool[f((e_h,r,e_t), (r_{hd},\mathbf{r_{b}}))], (r_{hd},\mathbf{r_{b}})\in\Psi_m,
\end{equation}
where \textit{Pool} denotes the pooling operator such as mean pooling or max pooling. To trace which facts should be deleted, we select triples in line with their ranks on influence values to the KG and control the removal number to the perturbation budget $\delta$.

\subsection{Adversarial Addition}   
In the process of addition attack, our objective is to insert noisy triples distorting the KG's structural patterns. These inserted triples should ideally possess minimal plausibility. To this end, we focus on logic rules with lower confidence, since the low-confident rules usually represent unreliable symbolic patterns and the semantics of triples grounded to such rules are inherently weak. To make added perturbation triples hold significantly different semantics to the global semantics of the KG, we further disrupt these low-confident rules to get negative rules, 
% These negative rules 
which are utilized to generate perturbations that are semantically corrupt the KG's global semantics.

To begin with, we select top-$n$ rules for each $r\in\mathcal{R}$ with the lowest confidence values from $\Psi$ to get a set of rules $\Psi_{n}$ that will be further destroyed. For a rule $(r_{hd}, \mathbf{r_b})\in\Psi_{n}$, we randomly select one predicate of $\mathbf{r_b}$ and replace it with another predicate according to the following heuristics. 
We note that entities connected by a certain relation are semantically constrained by certain types. For example, the head entity linked to the relation \textit{hasJob} belongs to the type of Person, while entities connected with \textit{locatedIn} belong to the type of Place. 
% Therefore, 
Randomly replacing a predicate may lead to meaningless rules. For example, given a positive rule $bornIn(X,Z)\wedge{locatedIn(Z,Y)}\rightarrow{bornIn(X,Y)}$, though replacing the predicate $locatedIn$ by $hasJob$ results in a negative rule $bornIn(X,Z)\wedge{hasJob(Z,Y)}\rightarrow{bornIn(X,Y)}$, it cannot be grounded into the KG with body triples since $Z$ cannot serve as a place and a person at the same time.   
To avoid such meaninglessness and guarantee enough perturbation triples which are inferred by the negative rules, we design a heuristic predicate rewriting strategy based on the correlation between different relations in the KG. 
We define the correlation value of a relation $r_i$ w.r.t. a given relation $r$ as: 
\begin{equation}
    cor(r\rightarrow r_i) = \frac{|\{e|e\in\mathcal{E}, r\in\mathcal{N}_{\mathcal{R}}(e)\wedge r_i\in\mathcal{N}_{\mathcal{R}}(e)\}|}{|\{e|e\in\mathcal{E}, r\in\mathcal{N}_{\mathcal{R}}(e)\}|},
\end{equation}
where $\mathcal{N}_{\mathcal{R}}(e)$ denotes the set of relations connecting entity $e$. Based on the correlation value, we can turn the existing logic rule $(r_{hd},\mathbf{r_b})$ into a negative rule $(r_{hd},\mathbf{r_b}')$ by replacing one relation $r_i$ in the relation sequence $\mathbf{r_b}$ by another relation $r_i'$ that has the highest correlation value w.r.t. $r_i$.

In this way, we can transform every single rule in $\Psi_{n}$ into a corresponding negative rule and obtain a collection of negative rules $\widehat{\Psi_{n}}$, which covers all relations in $\mathcal{G}$. Thus, we can generate diverse negative triples as they are inferred by employing rules in $\widehat{\Psi_{n}}$ over $\mathcal{G}$. Eventually, we make a sampling of all generated negative triples according to the relation distribution as final perturbation triples, whose amount is fixed to the perturbation budget $\delta$. 

\section{Experiments}
We implement experiments on 
% two datasets attacking 
seven typical KGE methods to address three concerns: \textbf{C1}- Can our proposed approaches effectively attack the untargeted scenarios? \textbf{C2} - What are the differences in the robustness of different KGE methods? \textbf{C3} - How does the implementation of logic rules influence the attacking performances? 

\subsection{Datasets and Evaluation Metrics}
Two widely-used datasets FB15k-237 \cite{toutanova-etal-2015-representing} and WN18RR \cite{dettmers2018convolutional} are employed in our experiments. FB15k-237 is a challenging subset of Freebase, which is a large-scale KG consisting of real-word facts, while WN18RR is derived from a large English lexical KG WordNet. The detailed statistics of these two datasets are listed in Table \ref{tab:dataset}. 

We carry out link prediction over all the triples in the test set. For each testing triple, we first generate negative triples by replacing the head or tail entity with all the other entities, then apply the KGE method to score the test triple and its corresponding negative triples, and finally sort them in the descending order of the score. We utilize MRR which calculates the mean reciprocal rank, and Hits@10, which is the proportion of testing triples that are ranked among the top-10, as the metrics. 
The greater degradation in these two metrics signifies a higher level of effectiveness of the attack.

\begin{table}[]
\setlength{\abovecaptionskip}{0.1cm}
\caption{The statistics of FB15k-237 and WN18RR.}
\resizebox{0.45\textwidth}{!}{
\begin{tabular}{cccccc}
\toprule
\textbf{Dataset} & \textbf{\#Entity} & \textbf{\#Relation} & \textbf{\#Train} & \textbf{\#Valid} & \textbf{\#Test} \\
\midrule
FB15k-237 & 14,541 & 237 & 272,115 & 17,535 & 20,466 \\
WN18RR & 40,943 & 11 & 86,835 & 3,034 & 3,134 \\
\bottomrule
\end{tabular}
}
\label{tab:dataset}
\vspace{-0.3cm}
\end{table}

\begin{table*}[]
\setlength{\abovecaptionskip}{0.1cm}
\caption{Performance ($\%$) of adversarial deletion under the perturbation ratio of $0.1$. Lower values indicate better attack results. \textbf{Bold} and \underline{underline} numbers denote optimal results and sub-optimal results, respectively. The \textcolor{red}{red} values in the subscripts denote the greatest relative drops on the Hits@10 metric.}
\resizebox{0.98\textwidth}{!}{%
\fontsize{40}{60}\selectfont
\begin{tabular}{cccclcclcclcclcclcclcclcc}
\toprule[3pt]
\multirow{3}{*}{\textbf{Dataset}} & \multirow{3}{*}{\textbf{Attacker}} & \multicolumn{8}{c}{\textbf{Fact-based KGE}} & \textbf{} & \multicolumn{5}{c}{\textbf{GNN-based KGE}} & \textbf{} & \multicolumn{5}{c}{\textbf{Rule-based KGE}} \\ 
\cmidrule{3-10} \cmidrule{12-16} \cmidrule{18-22} 
                                  &                                    & \multicolumn{2}{c}{\textbf{TransE}}         &           & \multicolumn{2}{c}{\textbf{DistMult}}      &           & \multicolumn{2}{c}{\textbf{ComplEx}} & & \multicolumn{2}{c}{\textbf{R-GCN}} &  & \multicolumn{2}{c}{\textbf{CompGCN}} & \textbf{} & \multicolumn{2}{c}{\textbf{RNNLogic}} &  & \multicolumn{2}{c}{\textbf{NCRL}} \\                                  
\cmidrule{3-4} \cmidrule{6-7} \cmidrule{9-10} \cmidrule{12-13} \cmidrule{15-16} \cmidrule{18-19} \cmidrule{21-22} 
 &              & \fontsize{30}{60}\textbf{MRR}    & \fontsize{30}{60}\textbf{Hits@10}  &  & \fontsize{30}{60}\textbf{MRR}     & \fontsize{30}{60}\textbf{Hits@10}   &  & \fontsize{30}{60}\textbf{MRR}     & \fontsize{30}{60}\textbf{Hits@10}  &  & \fontsize{30}{60}\textbf{MRR}   & \fontsize{30}{60}\textbf{Hits@10}  &  & \fontsize{30}{60}\textbf{MRR}    & \fontsize{30}{60}\textbf{Hits@10}   & \textbf{} & \fontsize{30}{60}\textbf{MRR}   & \fontsize{30}{60}\textbf{Hits@10}   &  & \fontsize{30}{60}\textbf{MRR}   & \fontsize{30}{60}\textbf{Hits@10} \\ 
\midrule
\multirow{5}{*}{\rotatebox{90}{\textbf{FB15k-237}}} & No Attack   & 29.36             & 48.20    & & 25.42             & 40.40    & & 27.04          & 43.04    & & 27.73    & 42.68                                                                &             & 35.15             & 53.30    & & 33.08             & 48.52    & & 40.68          & 54.12          \\ 
\cmidrule{2-22} 
                                                    & Random      & 26.81             & 45.44    & & 23.49             & 38.53    & & 25.07          & 41.01    & & 24.74    & 40.38           &             & \underline{30.42} & 49.42    & & 29.23      & 45.31    & & 42.16      & 53.99     \\
                                                    & CosAttack   & \textbf{26.68}    & 45.45    & & \underline{23.28} & 38.19    & & \textbf{24.85}    & \underline{40.77} & & 24.18 & 40.02  &             & \textbf{30.33}    & 49.33    & & \underline{28.46} & 44.25 & & 37.43       & 51.45           \\
                                                    & GradAttack  & 26.76             & \underline{45.38 } & & 23.44   & \underline{38.18} & & 24.92 & 40.93    & & \underline{24.06}           & 40.24           &  & \underline{30.42} & 49.37    & & 28.65      & \underline{43.99} & & 42.61      & 56.94      \\
\cmidrule{2-22} 
                                                    &\textbf{Ours}& 26.72             & \textbf{44.76}$_{\mathbf{\color{red}7\%\downarrow}}$ & & \textbf{23.06} &\textbf{37.45}$_{\mathbf{\color{red}7\%\downarrow}}$ & & \underline{24.90} & \textbf{40.15}$_{\mathbf{\color{red}7\%\downarrow}}$ &              & 25.21       & \textbf{ 39.65 }$_{\mathbf{\color{red}7\%\downarrow}}$  &  & 31.73     & \textbf{49.22}$_{\mathbf{\color{red}8\%\downarrow}}$  & & \textbf{25.01} & \textbf{38.09}$_{\mathbf{\color{red}21\%\downarrow}}$ & & \textbf{34.29} & \textbf{48.43}$_{\mathbf{\color{red}11\%\downarrow}}$   \\ 
\midrule[1.5pt]
\multirow{5}{*}{\rotatebox{90}{\textbf{WN18RR}}}    & No Attack   & 20.20             & 49.90     & & 39.31            & 52.94      & & 44.86          & 55.82    & & 37.45    & 42.49                                                               &             & 46.39            & 53.80     & &  46.44           & 52.49      & & 23.45          & 60.91            \\ 
\cmidrule{2-22} 
                                                    & Random      & \underline{18.33} & \underline{45.20} & & \underline{35.50} & \underline{48.28} & & \underline{40.06} & \underline{51.16}   &             & \underline{32.90} & \underline{38.88} & & \underline{40.48} & \underline{48.42} & & \underline{40.25} & \underline{46.62}   &             & 26.72             & 63.08           \\
                                                    & CosAttack   & 19.40             & 47.45     & & 37.00            & 50.64      & & 41.66       & 52.86     & & 34.14    & 40.16           &             & 41.47             & 49.76     & & 41.75            & 49.43      & & 26.16       & 61.23           \\
                                                    & GradAttack  & 18.84             & 46.51     & & 36.00            & 50.00      & & 40.76       & 52.76     & & 33.09    & 39.58            &             & 40.85             & 49.49     && 40.52             & 47.78      & & 24.47       & 61.74    \\ 
\cmidrule{2-22} 
                                                    &\textbf{Ours}& \textbf{17.89}    & \textbf{42.39}$_{\mathbf{\color{red}15\%\downarrow}}$    & & \textbf{33.44}    & \textbf{44.94}$_{\mathbf{\color{red}15\%\downarrow}}$    & & \textbf{38.67}    & \textbf{47.42}$_{\mathbf{\color{red}15\%\downarrow}}$     &             &\textbf{30.36}     & \textbf{ 35.53}$_{\mathbf{\color{red}16\%\downarrow}}$   & & \textbf{37.62}    & \textbf{44.31}$_{\mathbf{\color{red}18\%\downarrow}}$    & &  \textbf{32.88}   &  \textbf{36.34}$_{\mathbf{\color{red}31\%\downarrow}}$    &             & \textbf{24.03}             & \textbf{61.07}           \\ 
\bottomrule[3pt]
\end{tabular}
}
\label{tab:main1}
\end{table*}

\begin{table*}[]
\setlength{\abovecaptionskip}{0.1cm}
\caption{Performance ($\%$) of adversarial addition under the perturbation ratio of $0.1$.  Its settings are consistent with Table 2.}
\resizebox{0.98\textwidth}{!}{%
\fontsize{40}{60}\selectfont
\begin{tabular}{cccclcclcclcclcclcclcclcc}
\toprule[3pt]
\multirow{3}{*}{\textbf{Dataset}} & \multirow{3}{*}{\textbf{Attacker}} & \multicolumn{8}{c}{\textbf{Fact-based KGE}} & \textbf{} & \multicolumn{5}{c}{\textbf{GNN-based KGE}} & \textbf{} & \multicolumn{5}{c}{\textbf{Rule-based KGE}} \\ 
\cmidrule{3-10} \cmidrule{12-16} \cmidrule{18-22} 
                                  &                                    & \multicolumn{2}{c}{\textbf{TransE}}         &           & \multicolumn{2}{c}{\textbf{DistMult}}      &           & \multicolumn{2}{c}{\textbf{ComplEx}} & & \multicolumn{2}{c}{\textbf{R-GCN}} &  & \multicolumn{2}{c}{\textbf{CompGCN}} & \textbf{} & \multicolumn{2}{c}{\textbf{RNNLogic}} &  & \multicolumn{2}{c}{\textbf{NCRL}} \\                                  
\cmidrule{3-4} \cmidrule{6-7} \cmidrule{9-10} \cmidrule{12-13} \cmidrule{15-16} \cmidrule{18-19} \cmidrule{21-22} 
 &              & \fontsize{30}{60}\textbf{MRR}    & \fontsize{30}{60}\textbf{Hits@10}  &  & \fontsize{30}{60}\textbf{MRR}     & \fontsize{30}{60}\textbf{Hits@10}   &  & \fontsize{30}{60}\textbf{MRR}     & \fontsize{30}{60}\textbf{Hits@10}  &  & \fontsize{30}{60}\textbf{MRR}   & \fontsize{30}{60}\textbf{Hits@10}  &  & \fontsize{30}{60}\textbf{MRR}    & \fontsize{30}{60}\textbf{Hits@10}   & \textbf{} & \fontsize{30}{60}\textbf{MRR}   & \fontsize{30}{60}\textbf{Hits@10}   &  & \fontsize{30}{60}\textbf{MRR}   & \fontsize{30}{60}\textbf{Hits@10} \\ 
\midrule
\multirow{6}{*}{\rotatebox{90}{\textbf{FB15k-237}}}  & No Attack      & 29.36           & 48.20       & & 25.42            & 40.40              & & 27.04            & 43.04                    
                                                     &                & 27.73           & 42.68       & & 35.15            & 53.30              & & 33.08            & 48.52                    &                & 40.68           & 54.12               \\ 
\cmidrule{2-22} 
                                                     & Random     & \underline{24.82} & \underline{44.30} & & 25.70        & 40.89              & & 27.62            & 43.71                    &                & 25.52           & \textbf{39.79}$_{\mathbf{\color{red}7\%\downarrow}}$  & & 34.61        & 52.62              & & 33.22            & 47.93                   &                & 36.21           & \underline{49.85}   \\
                                                     & CosAttack      & 26.83           & 44.70       & & \textbf{23.64}   & 38.68              & &\underline{25.57} & \underline{41.59}        &                & 26.36           & 41.00       & &\underline{34.19} &\underline{52.23}   & &\textbf{32.30}& \textbf{47.44}$_{\mathbf{\color{red}2\%\downarrow}}$         &                & 35.95           & \textbf{49.16}      \\
                                                     & GradAttack     & 26.66           & 44.38       & &\underline{23.77} & \underline{38.50}  & & \textbf{25.54}   & 41.63                    &                & \textbf{25.42}  & 40.48       & & \textbf{34.11}   & \textbf{52.08}$_{\mathbf{\color{red}2\%\downarrow}}$     & & 32.55            & 48.08                   &                & \underline{35.46}       & 51.32       \\
                                                     & ComAttack      & 26.98           & 45.87       & & 25.04            & 40.01              & & 27.02            & 42.98                   &                & 27.00           & 42.09       & & 35.01            & 53.18              & & 35.64            & 51.67                   &                & 36.06           & 51.88               \\ 
\cmidrule{2-22} 
                                                     & \textbf{Ours}  & \textbf{24.77}  & \textbf{42.25}$_{\mathbf{\color{red}12\%\downarrow}}$  & & 23.90        & \textbf{38.26}$_{\mathbf{\color{red}5\%\downarrow}}$     & & 25.87            & \textbf{41.27}$_{\mathbf{\color{red}4\%\downarrow}}$           &            & \underline{25.44} & \underline{40.34} & & 34.84        & 53.03              & & \underline{32.40}           & \underline{47.46}                   &                & \textbf{34.35} & \textbf{49.16}$_{\mathbf{\color{red}9\%\downarrow}}$       \\ 
\midrule[1.5pt]
\multirow{6}{*}{\rotatebox{90}{\textbf{WN18RR}}}    & No Attack       & 20.20           & 49.90       & & 39.31            & 52.94              & & 44.86            & 55.82                   
                                                    &                 & 37.45           & 42.49       & & 46.39            & 53.80              & &  46.44           & 52.49                   &                 & 23.45           & 60.91            \\ 
\cmidrule{2-22} 
                                                    & Random         & 18.67            & 47.13       & & 39.59            & 51.55              & & 44.01            & 54.48                   &                & 35.89            & 41.80       & &\underline{44.41} & \underline{51.61}  & & 44.13            & 50.03                   &                & 21.40            & 53.00            \\
                                                    & CosAttack      & 18.67            & 44.51       & & 38.54            & 51.26              & & 43.45            & 54.23                   &            & \underline{35.53} & \underline{41.50} & & 44.47         & 51.96              & & 40.92            & 46.33                   &            & \underline{19.65}    & 48.98            \\
                                                    & GradAttack & \underline{15.72} & \underline{44.48} & & 38.19         & \underline{51.20}  & & 44.87            & \underline{53.86}        &            & \textbf{ 35.19}   & \textbf{41.40}$_{\mathbf{\color{red}3\%\downarrow}}$ & & \textbf{44.07}   & \textbf{51.50}$_{\mathbf{\color{red}4\%\downarrow}}$     & &\underline{40.92} & \underline{46.83}                   &                 & 19.76           & \underline{45.09}      \\
                                                    & ComAttack       & 20.27           & 48.85       & &\underline{33.19} & 51.50              & &\underline{36.59} & 54.56                   &                 & 36.04           & 42.36       & & 46.19            & 53.88              & & 45.55            & 52.05                   &                 & 22.05           & 50.03            \\ 
\cmidrule{2-22} 
                                                    & \textbf{Ours}   & \textbf{15.36}  &\textbf{43.17}$_{\mathbf{\color{red}13\%\downarrow}}$ & & \textbf{24.43} & \textbf{47.65}$_{\mathbf{\color{red}10\%\downarrow}}$     & & \textbf{33.32}   & \textbf{52.41}$_{\mathbf{\color{red}6\%\downarrow}}$          &                 & 37.59           & 42.69       & & 45.77            & 52.86              & & \textbf{39.39}   & \textbf{42.73}$_{\mathbf{\color{red}19\%\downarrow}}$          &                 & \textbf{18.57} & \textbf{43.65}$_{\mathbf{\color{red}28\%\downarrow}}$   \\ 
\bottomrule[3pt]
\end{tabular}
}
\label{tab:main2}
\end{table*}

\subsection{Evaluated KGE Methods}
We conduct experiments on the impact on link prediction performance across seven typical KGE methods before and after attacks. They can be broadly classified into three categories:

\noindent $\bullet$ \textbf{Fact-based KGE methods}: This kind of method aims at learning entity and relation embeddings, and assessment of the plausibility of individual facts. Within this study, we include TransE~\cite{bordes_2013}, which is well-known as the most representative KGE method, along with DistMult~\cite{2014Embedding} and its extension model ComplEx~\cite{pmlr-v48-trouillon16}, which projects entities and relations into a complex space.

\noindent $\bullet$ \textbf{GNN-based KGE methods}: Methods based on GNNs leverage the topological structure of KGs through a message-passing mechanism facilitated by GNNs. We evaluate RGCN \cite{10.1007/978-3-319-93417-4_38} and CompGCN \cite{Vashishth2020Composition-based}, both of which incorporate relational types into the GNN updating process. 

\noindent $\bullet$ \textbf{Rule-based KGEs}: These methods learn logic rules from the KG and apply them for inferring missing triples. In our benchmarks, We select RNNLogic \cite{qu2021rnnlogic} and NCRL \cite{cheng2023neural}. 

\subsection{Baselines and Implementation Details}
Considering there are no existing works for untargeted attacks, we compare our proposed attacks to random modification and several state-of-the-art targeted attackers by adapting them to the untargeted setting: 

\noindent $\bullet$ \textbf{Random}: This method randomly selects triples in the training set to delete, and adds negative triples by randomly replacing the head entity or tail entity of the original training triple. 

\noindent $\bullet$ \textbf{CosAttack} \cite{DBLP:conf/emnlp/BhardwajKCO21} leverages the cosine similarity between the perturbation and the target triple, calculated with the triple-level representation. Given a target triple, CosAttack either deletes training triples that exhibit the highest cosine similarity or introduces noisy triples that are not similar to the target. 

\noindent $\bullet$ \textbf{GradAttack} \cite{DBLP:conf/emnlp/BhardwajKCO21} calculates the perturbations based on the gradient similarities. Similar to CosAttack, GradAttack removes triples with high gradient similarities or adds perturbation triples with diminished gradient similarity values. 

\noindent $\bullet$ \textbf{ComAttack} \cite{DBLP:conf/acl/BhardwajKCO20} only consider the adversarial addition setting. Close to the logic rules which are employed in this paper, ComAttack utilizes compositional relation patterns. It injects perturbation triples that align with the composition pattern, inducing KGE methods to predict decoy triples instead of given target triples.  

To implement these targeted attack methods without direct access to test triples, we randomly select a subset of the original training triples as the targets. 
% Besides, considering 
For another restriction of targeted attack methods that rely on the full knowledge of certain KGE methods, we apply TransE as the foundational KGE method for them.
We set the perturbation budget $\delta=\gamma\cdot|\mathcal{T}|$, where $\gamma\in\{0.05,0.10,0.15,0.20,0.25\}$ is the perturbation ratio. In our proposed attacks, we utilize prior logic rules with the length $L=2$. We set the $m=50$ and $n=10$ for selecting rules with the highest and lowest confidence for adversarial deletion and addition, respectively.

\subsection{Attack Effectiveness (\textbf{C1})}
\subsubsection{Overall Performance} We evaluate the attacks in both adversarial deletion and addition scenarios. The experimental results are listed in Table \ref{tab:main1} and Table \ref{tab:main2}. 

In adversarial deletion, our method outperforms the baselines over the majority of KGE models. 
Remarkably, on WN18RR, simply adapting targeted attacks such as CosAttack and GradAttack can only yield a marginal diminution. These targeted attacks even exhibit weaker performance than Random modifications, highlighting the obvious limitations of targeted attacks. 
%Compared to them, our method demonstrates significant superiority. 
Our deletion attack exerts the most substantial impact when applied to RNNLogic on WN18RR, inducing a decline of $20.97\%$ in MRR and a notable $30.76\%$ drop in Hits@10. On FB15k-237, our method also shows competitive performances in achieving the greatest Hits@10 reduction across all the KGE methods. 

In adversarial addition, our method also demonstrates a strong capability in attacking the fact-based KGE methods. For the GNN-based KGE methods, though our method does not always yield the best results, its performance is competitive to baseline attackers. This may be because GNNs learn more robust representations (see KGE robustness in Section 3.5).
As indicated in the last column of Table 3, our attack, which leverages logic rules, effectively hinders NCRL from learning essential rules, where NCRL's performance drops $28.34\%$ in Hits@10 on WN18RR.

We also observe higher performance drops on WN18RR than on FB15k-237. In scenarios involving both adversarial deletion and addition attacks, KGE methods suffer a relatively larger drop in link prediction results on WN18RR than on FB15k-237. This may be attributed primarily to the sparser graph structure of WN18RR, which is more susceptible to data biases under adversarial attacks. 

\begin{table}[]
\setlength{\abovecaptionskip}{0.1cm}
\caption{Performances of TransE on highly ranked targets.}
\begin{tabular}{cccccc}
\toprule
\multirow{2}{*}{\textbf{Attacker}} & \multicolumn{2}{c}{\textbf{FB15k-237}} &           & \multicolumn{2}{c}{\textbf{WN18RR}} \\ 
\cline{2-3} \cline{5-6} 
                                  & \textbf{MRR}      & \textbf{Hits@1}    & \textbf{} & \textbf{MRR}     & \textbf{Hits@1}  \\ 
\midrule
No Attacker                 & 51.59             & 36.74              &           & 41.51            & 2.45             \\ 
\midrule
ComAttack               & 49.31             & 33.41              &           & 39.68            & 0.58             \\
\textbf{Ours}                     & \textbf{47.10}    & \textbf{32.44}     & \textbf{} & \textbf{37.01}   & \textbf{0.37}    \\ 
\bottomrule
\end{tabular}
\label{tab:highRank}
\vspace{-0.4cm}
\end{table}

\begin{figure}[]
  \centering
  \subfloat[Results on FB15k-237]
    {
        \includegraphics[width=0.225\textwidth]{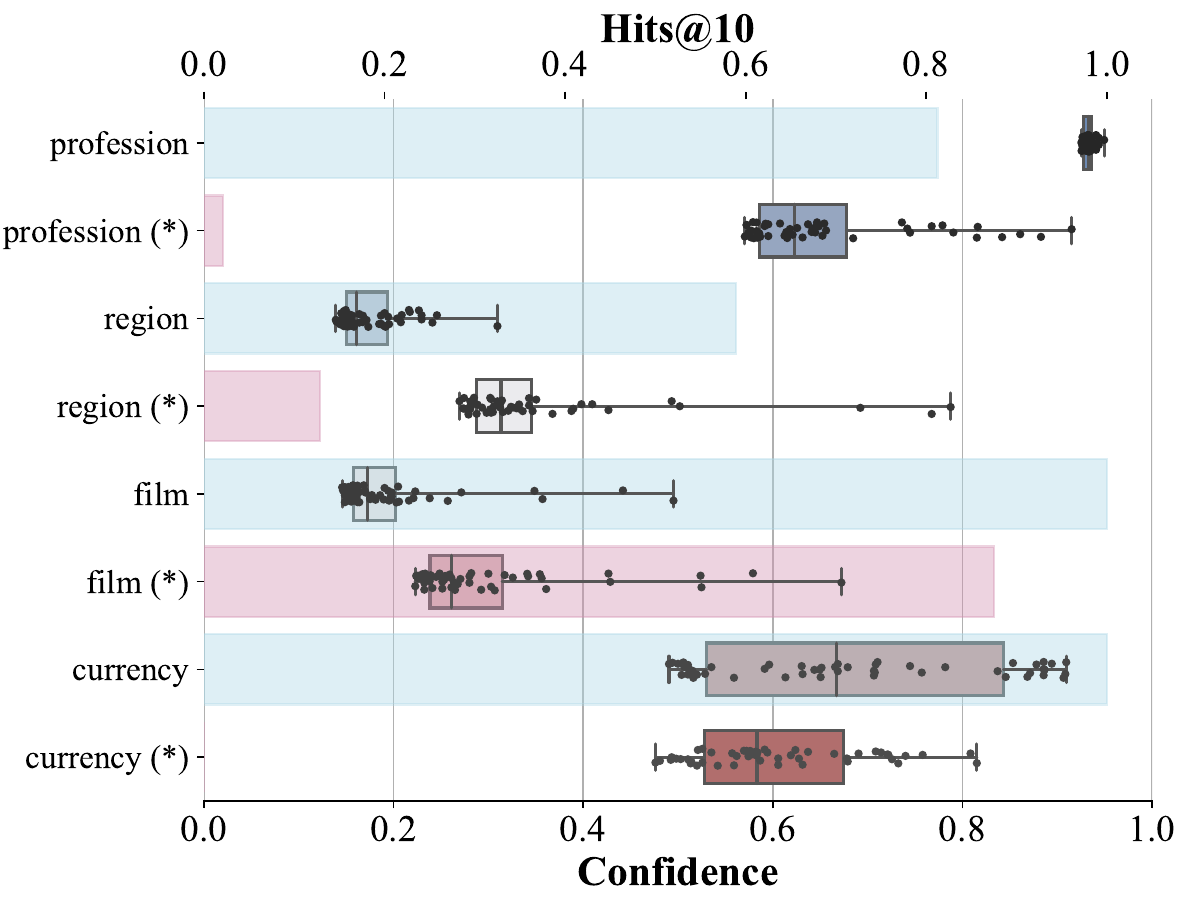}
    }
    \subfloat[Results on WN18RR]
    {
        \includegraphics[width=0.225\textwidth]{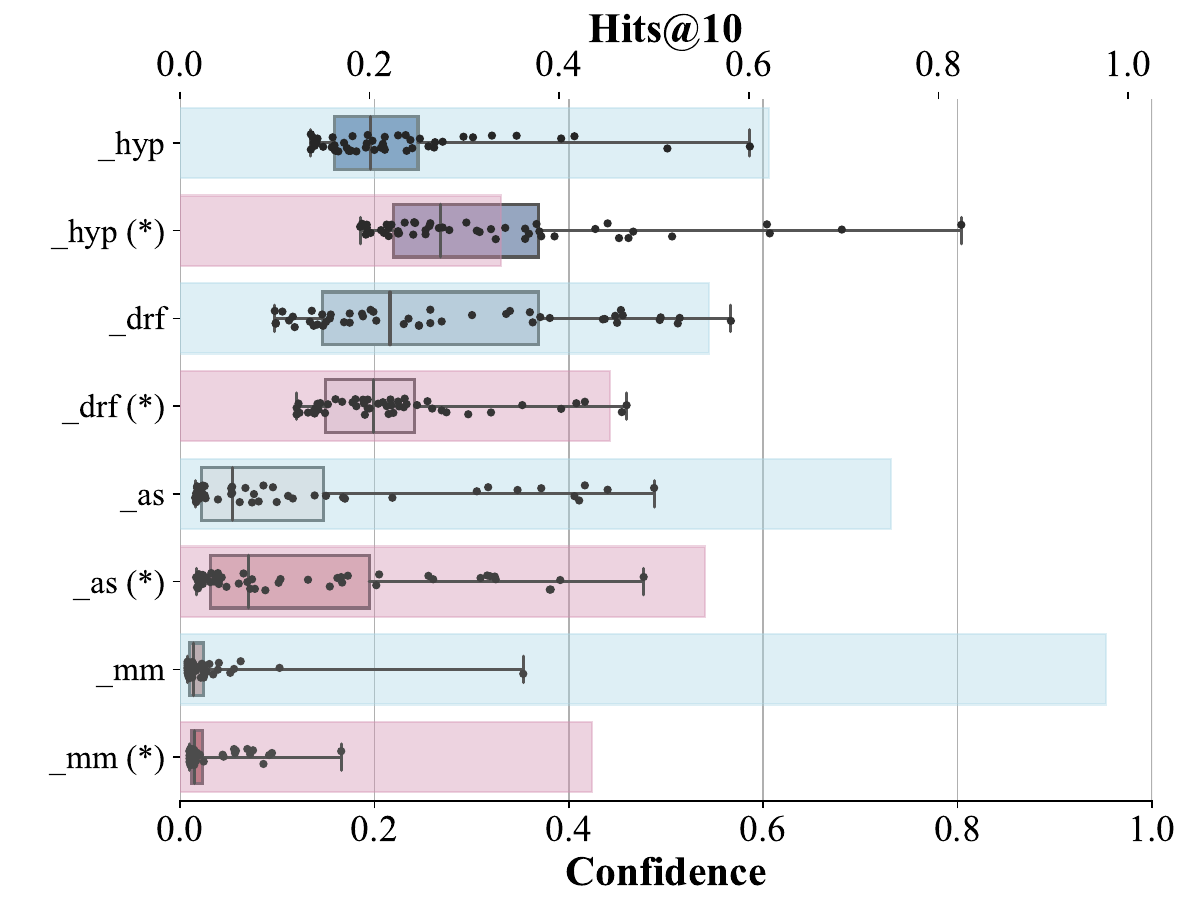}
    }
  \setlength{\abovecaptionskip}{0.1cm}
  \caption{The confidence distribution of different rule bodies and link prediction results of NCRL on different relations. ($\ast$) denotes the results under an addition attack. The hierarchy structure of the relations in FB15k-237 are simplified to its final split in (a); the relations \textit{\_derivationally\_related\_form}, \textit{\_hypernym}, \textit{\_also\_see} and \textit{\_member\_meronym} are marked as \_drf, \_hyp, \_as and \_mm in (b) for convenience.
  }\label{fig:rule}
  \Description{}
  \vspace{-0.4cm}
\end{figure}

\subsubsection{Attack Performance on Highly Ranked Triples.}
We extend our proposed attacks with the focus on degrading high-ranked test triples, which are commonly selected as target triples in previous targeted attack works\cite{ijcai2019p674,DBLP:conf/emnlp/BhardwajKCO21}. Motivated by \cite{DBLP:conf/acl/BhardwajKCO20,DBLP:conf/emnlp/BhardwajKCO21}, we select test triples with $ranks \leq 10$ by a certain KGE model and set MRR and Hits@1 as the evaluation metrics. We mainly compare with ComAttack, which considers the composition inference pattern similar to logic rules. For a fair comparison, we select TransE as the basic KGE model and constrain the perturbation budget of our methods equal to the perturbation number in ComAttack based on TransE. From the data presented in Table \ref{tab:highRank}, it is evident that although our proposed attack is primarily designed to impair the overall link prediction capabilities of KGE models, it also exhibits competitive effectiveness in impacting the results over triples that KGE methods predict can easily predict.

\begin{figure}[h]
  \centering
  \subfloat[FB15K-237]
    {
        \includegraphics[width=0.225\textwidth]{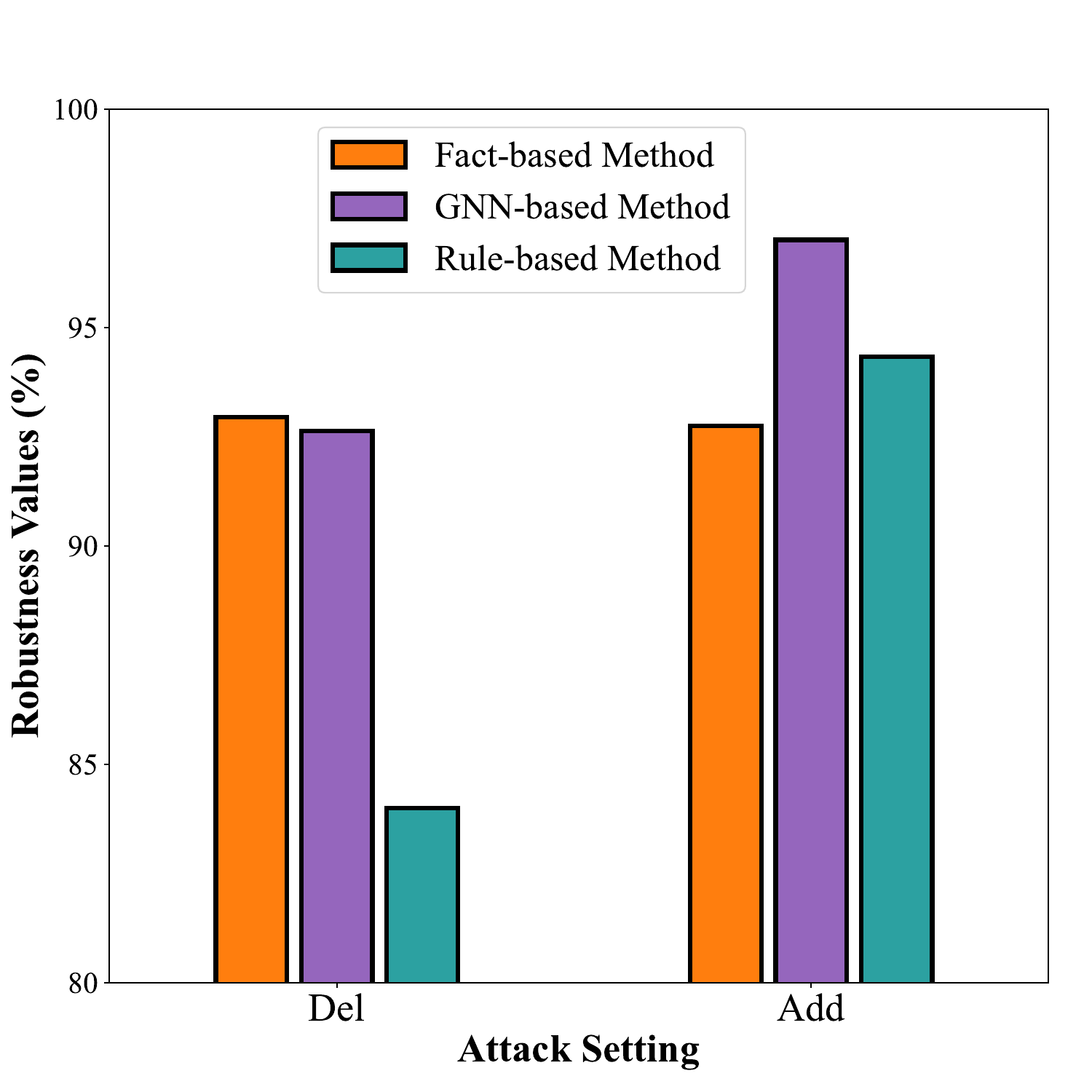}
    }
    \subfloat[WN18RR]
    {
        \includegraphics[width=0.225\textwidth]{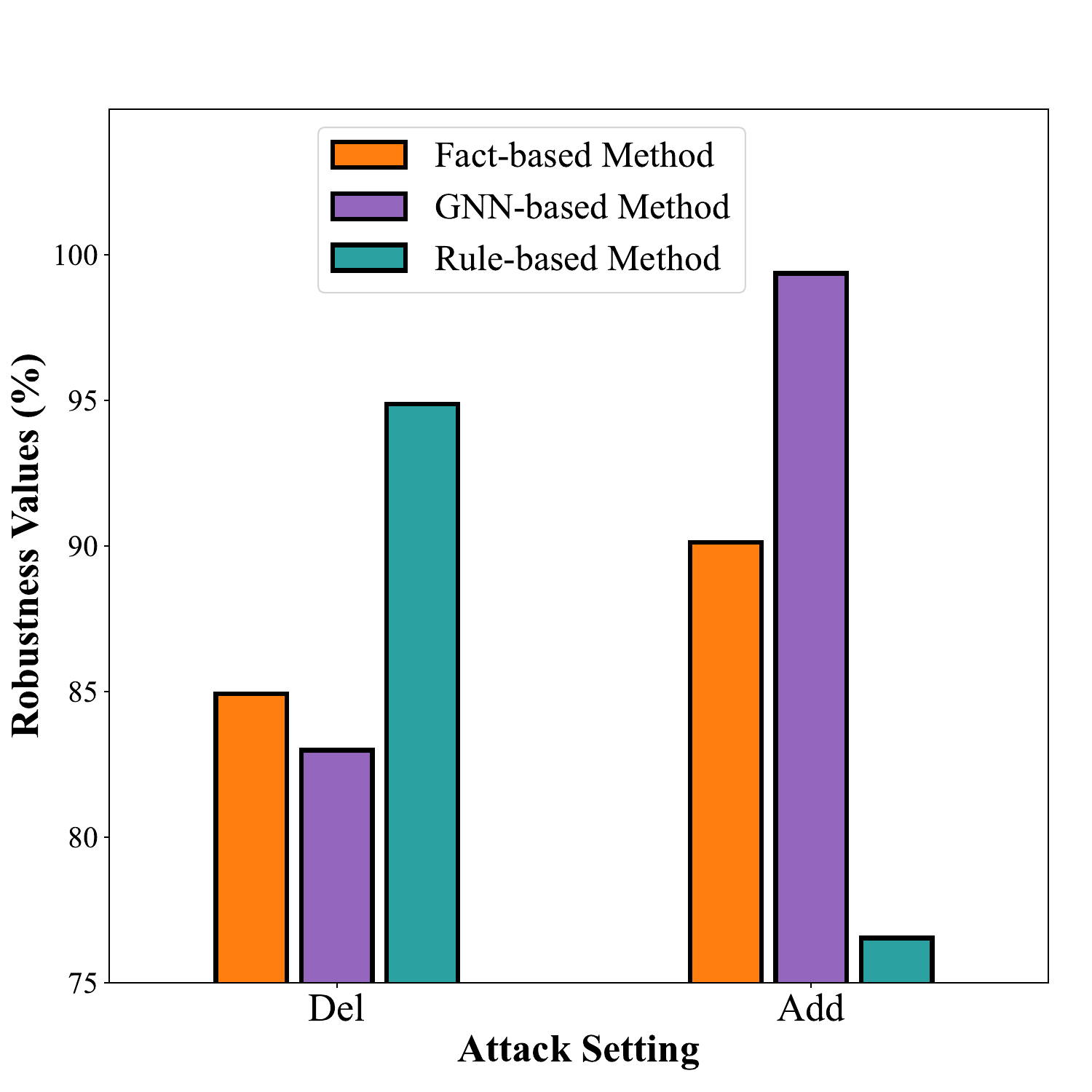}
    }
  \setlength{\abovecaptionskip}{0.1cm}
  \caption{The comparison of robustness between different classes of KGE methods. }\label{fig:comparison}
  \Description{}
  \vspace{-0.6cm}
\end{figure}

\subsubsection{Impact on the Rules}
Since our method uses rules to create the attacks, we compare the rules extracted before and after attacks. 
We conducted a statistical analysis of the confidence distribution of the extracted rules with different rule heads, and we also examined the attack impact on the link prediction results of these head relations.
For each head, we collect 50 rules with the highest confidence values and visualize the distribution in Figure \ref{fig:rule}. 
We find that when the valid rules are violated, the link prediction results reduce correspondingly. For example, the adversarial addition attack disrupts the logic rules on \textit{profession} in the FB15k-237 dataset, and the Hits@10 value on this relation reduces dramatically. As for the relation \textit{region} in FB15k-237 and \textit{\_hypernym} in WN18RR, the decrease in link prediction results may be caused by extracting many high-confident negative rules. In general, our addition attack demonstrates the great ability to insert perturbation symbolic patterns to KGs, which induce KGE methods capturing negative logic rules rather than positive ones.

\subsection{KGE Robustness (\textbf{C2})}
\subsubsection{Comparison in KGE Robustness.}
For a direct comparison of the robustness of different classes of KGE methods, we computed the robustness value, which is the division of Hits@10 values between attacked and non-attack results, of each KGE method and assigned the average values for their category of KGE methods. As shown in Figure \ref{fig:comparison}, on FB15k-237, Facted-based KGE methods show great robustness to the deletion attack, while the robustness of Rule-based methods is weak. And all three kinds of KGE methods are competitively robust to the adversarial addition on FB15k-237. Interestingly, Rule-based methods generally obtain superior robustness than the other two classes of KGE methods against the deletion attack on WN18RR. However, they also exhibit the poorest robustness to the adversarial addition on WN18RR.

\begin{figure*}[htbp]
    \centering
    \subfloat[TransE on FB15k-237]
    {
        \includegraphics[width=0.225\textwidth]{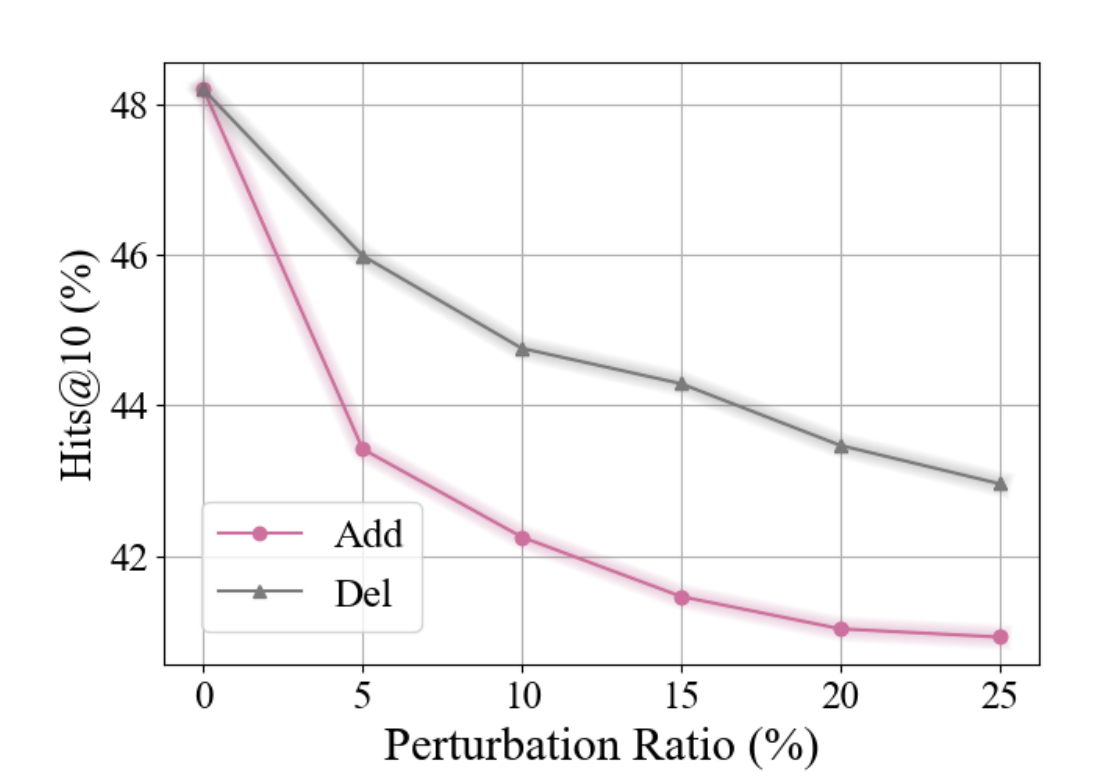}
    }
    \subfloat[DistMult on FB15k-237]
    {
        \includegraphics[width=0.225\textwidth]{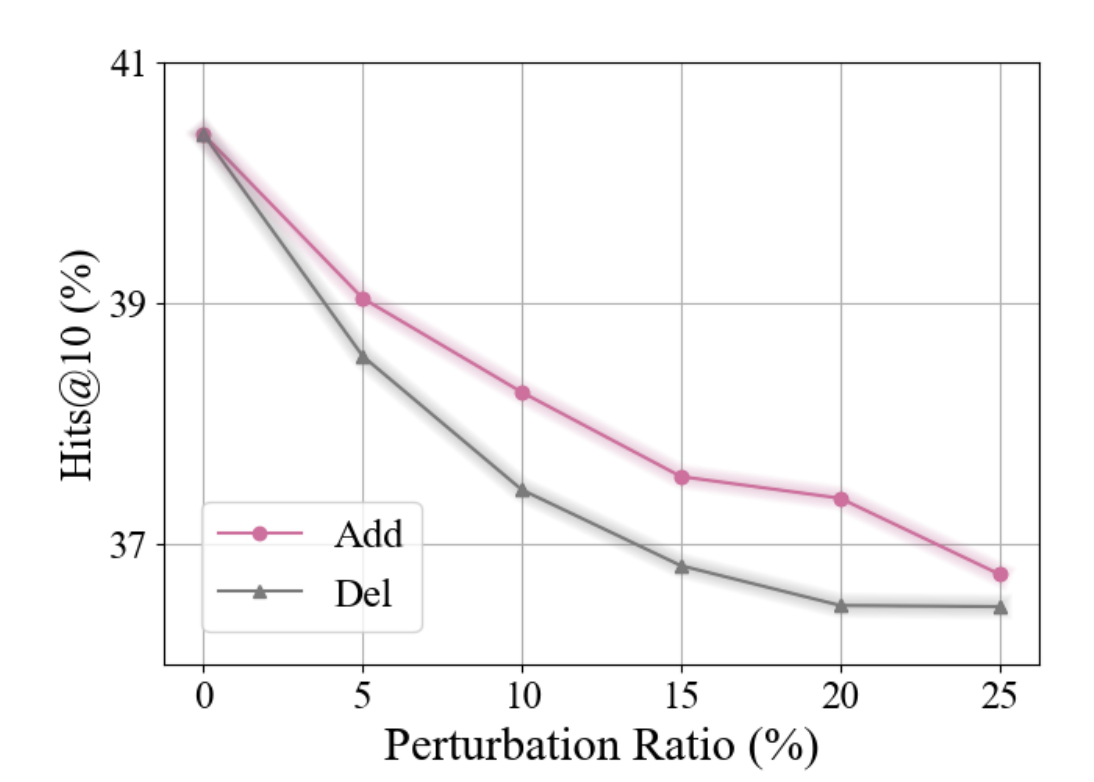}
    }
    \subfloat[RGCN on FB15k-237]
    {
        \includegraphics[width=0.225\textwidth]{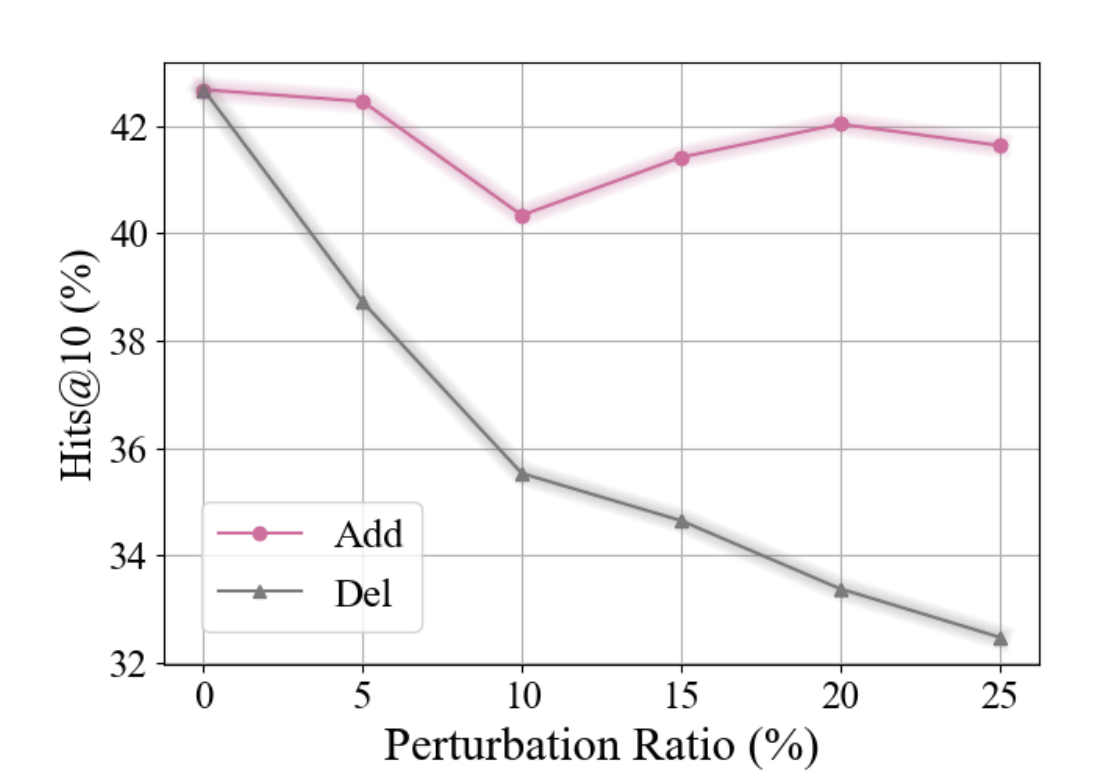}
    }
    \subfloat[RNNLogic on FB15k-237]
    {
        \includegraphics[width=0.225\textwidth]{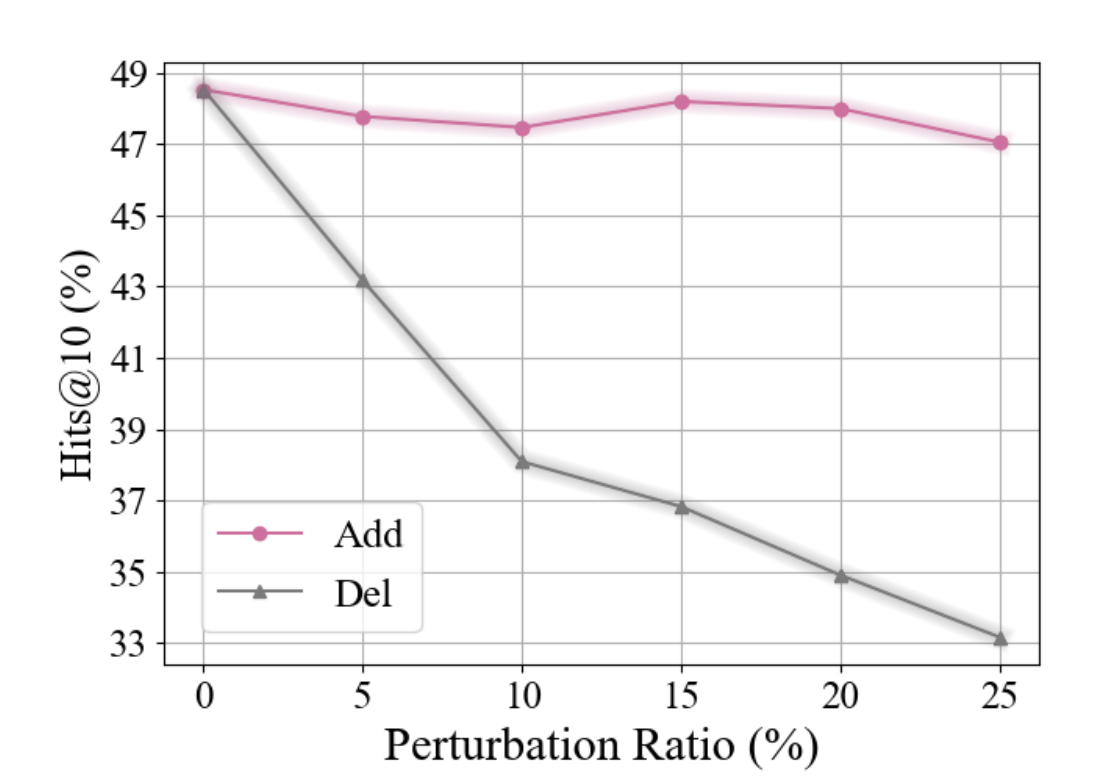}
    } \\
    \vspace{-0.4cm}
    \subfloat[TransE on WN18RR]
    {
        \includegraphics[width=0.225\textwidth]{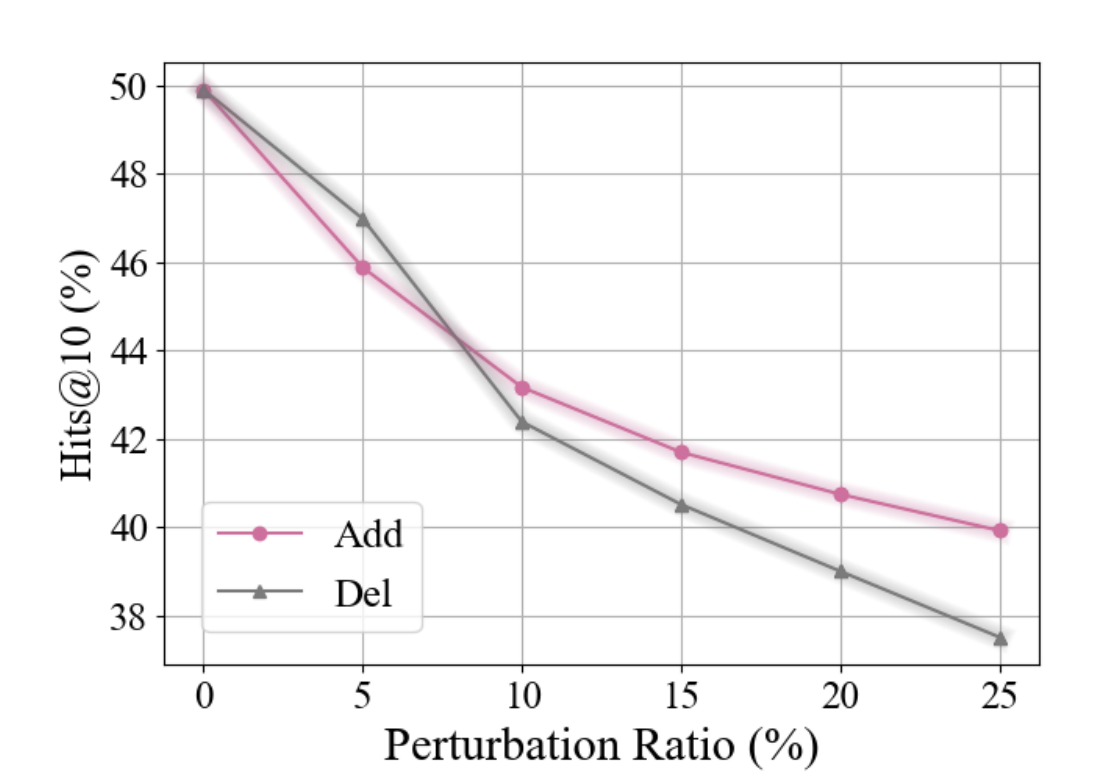}
    }
    \subfloat[DistMult on WN18RR]
    {
        \includegraphics[width=0.225\textwidth]{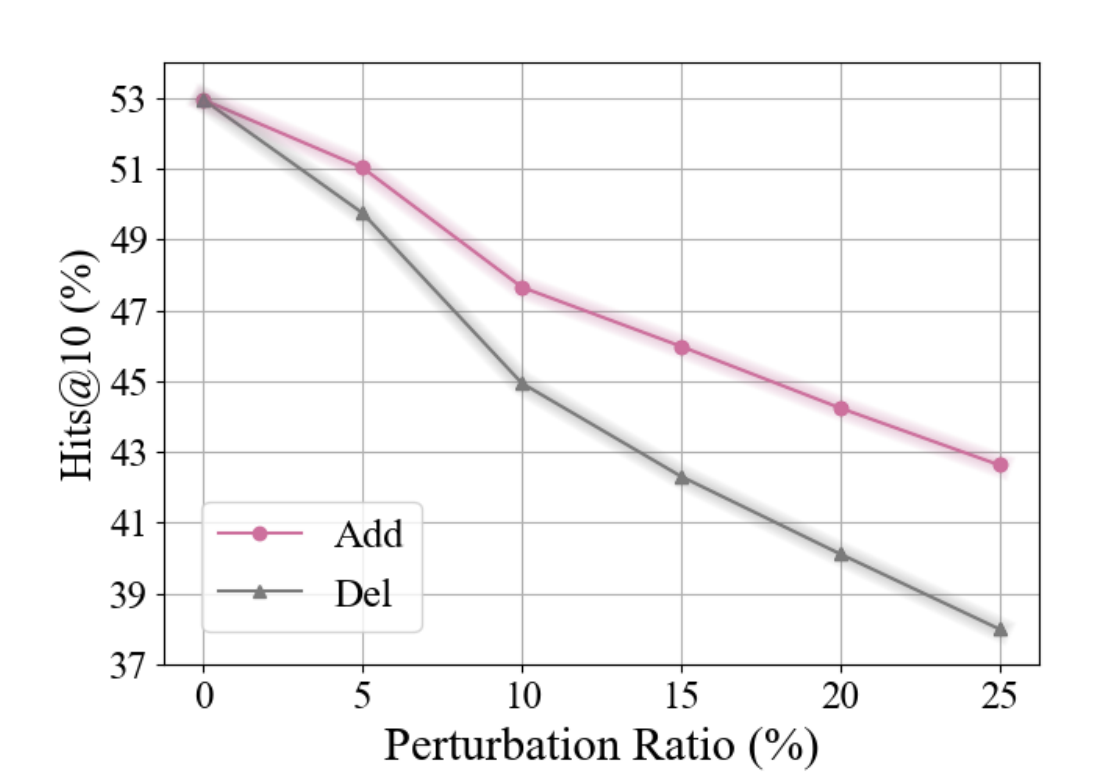}
    }
    \subfloat[RGCN on WN18RR]
    {
        \includegraphics[width=0.225\textwidth]{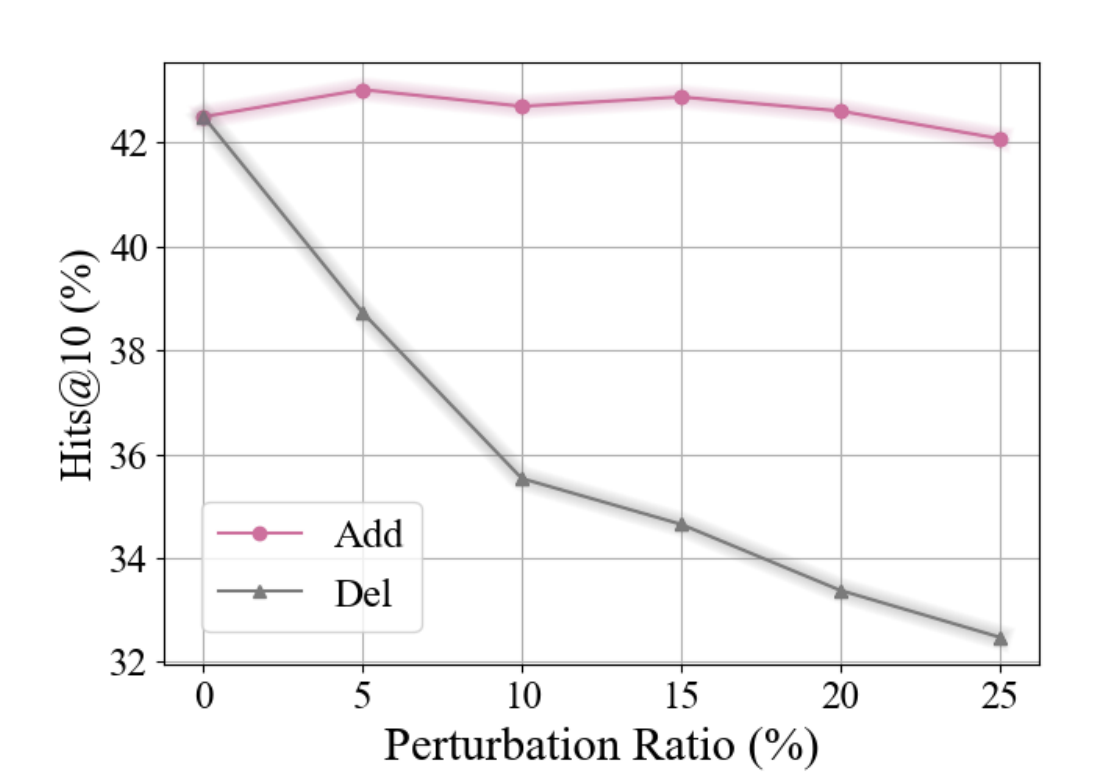}
    }
    \subfloat[RNNLogic on WN18RR]
    {
        \includegraphics[width=0.225\textwidth]{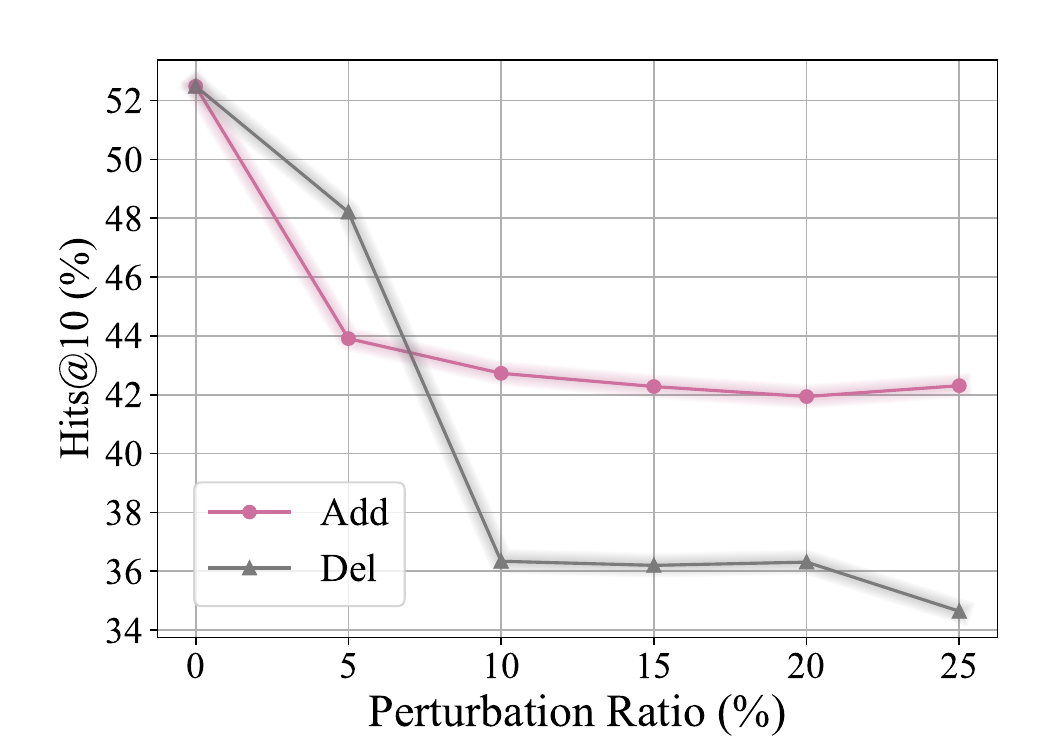}
    }
    \setlength{\abovecaptionskip}{0.1cm}
    \caption{Link prediction performances of KGEs under different perturbation ratios.}
    \label{fig:ratio}
\end{figure*}

\subsubsection{Robustness to Growing Perturbation Budget}
Besides the general comparison, we also focus on the robustness of a single KGE method to two different types of attacks under different perturbation budgets. As shown in Figure \ref{fig:ratio}, generally, KGE methods are more robust to adversarial additions than to adversarial deletions, where RGCN demonstrates the strongest robustness against addition attacks on two datasets. As the number of deleted triples grows, the performance of RGCN drops sharply. It indicates the robustness of GNN-based methods relies heavily on the density of the graph structure, which helps GNNs learn useful topology patterns. Note that TransE is more vulnerable to added perturbations on FB15k-237 as shown in Figure \ref{fig:ratio} (a), which illustrates that TransE is easily affected by the injected noise and learns biased KG presentations. Interestingly, on WN18RR, although adding a small proportion of perturbed triples obviously decreases the performance of RNNLogic. RNNLogic is persistent with the growth of added perturbations.

\subsubsection{Analyses of KG Embeddings}
\begin{figure}[]
    \centering
    \includegraphics[width=0.45\textwidth]{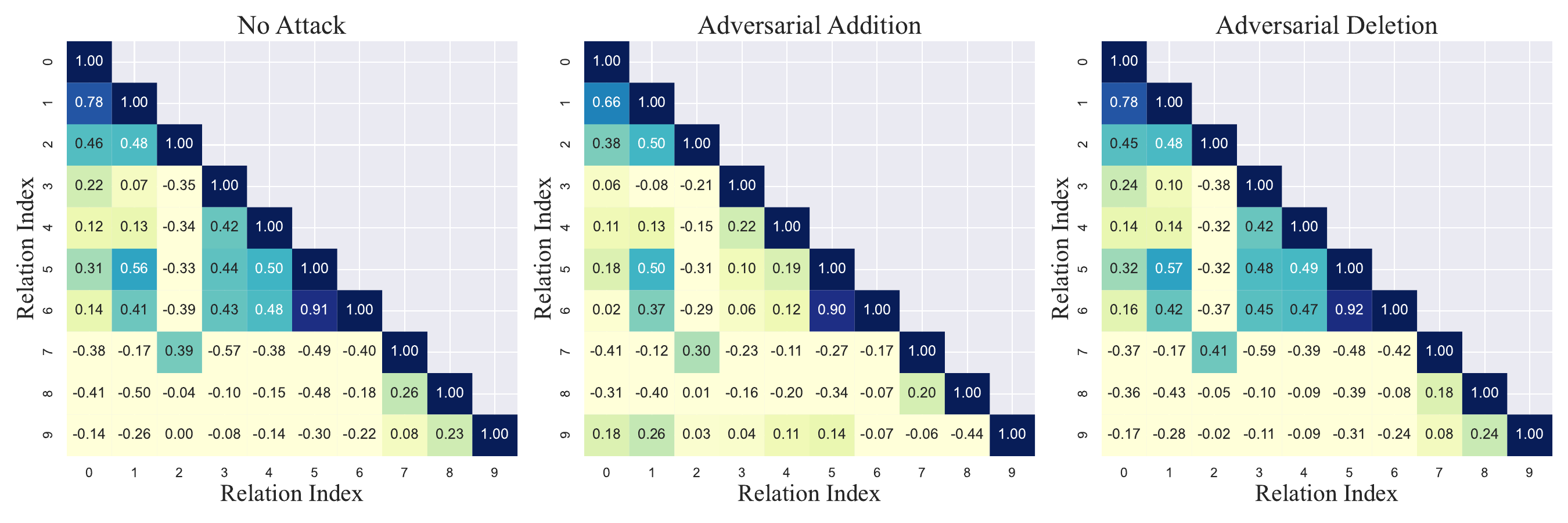}
    \setlength{\abovecaptionskip}{0.1cm}
    \caption{Relation similarity visualization on FB15k-237. Selected relations are generally categorized to \slash film\slash film\slash, \slash people\slash person\slash, \slash music\slash genre\slash and \slash location\slash location\slash.}\label{fig:visiual}
\end{figure}

As mentioned above, TransE shows weaker robustness to the addition attack than the deletion attack on FB15k-237, which is quite different from many other KGE methods like DistMult. To further figure out how attacks lead to such results, we explore changes of specific embeddings learned by TransE before and after attacks.  
From Figure \ref{fig:visiual}, it can be seen that TransE can well distinguish the different relations without adversarial attacks, and it can get high similarity values within the relations of the same type. Similar results can also be found in the deletion setting, while embeddings of TransE after the addition attack fail to identify relations belonging to the same type, where the similarity scores decrease obviously. This well explains the tendency in Figure \ref{fig:ratio} (a) and implies an idea to confuse similar relations for adversarial attacks. Besides, this analysis suggests the importance of distinguishing relations of different types for improving the robustness of KGE methods.

\begin{table}[]
\setlength{\abovecaptionskip}{0.1cm}
\caption{Link prediction results on Hits@10($\%$) of employing different prior rule extractors.}
\resizebox{0.45\textwidth}{!}{
\begin{tabular}{cccclcclcc}
\toprule
\multirow{2}{*}{\textbf{Dataset}}            &  \multirow{2}{*}{\textbf{Extractor}}                   & \multicolumn{2}{c}{\textbf{TransE}}      & & \multicolumn{2}{c}{\textbf{RNNLogic}}      &           & \multicolumn{2}{c}{\textbf{NCRL}} \\
\cmidrule{3-4} \cmidrule{6-7} \cmidrule{9-10}
 &          & Del    & Add  &  & Del     & Add   &  & Del     & Add \\
\midrule
\multirow{2}{*}{\textbf{FB15k-237}} & RNNLogic   & \textbf{42.76}        & 43.87    & & \textbf{37.39}             & 48.68    & & 56.11          & \textbf{47.94} \\
                                    & NCRL       & 44.76        & \textbf{42.25}    & & 38.09             & \textbf{42.73}    & & \textbf{48.43}          & 49.16 \\
\midrule
\multirow{2}{*}{\textbf{WN18RR}}    & RNNLogic   & 46.90        & 45.39    & & 36.63             & 44.38    & & 63.24          & 48.02 \\
                                    & NCRL       & \textbf{42.39}        & \textbf{43.17}    & & \textbf{36.34}             & \textbf{42.73}    & & \textbf{61.07}          & \textbf{43.65} \\
\bottomrule
\end{tabular}
}
\label{tab:extractor}
\vspace{-0.4cm}
\end{table}

\begin{figure}[h]
\vspace{-0.5cm}
  \centering
  \subfloat[TransE]
    {
        \includegraphics[width=0.225\textwidth]{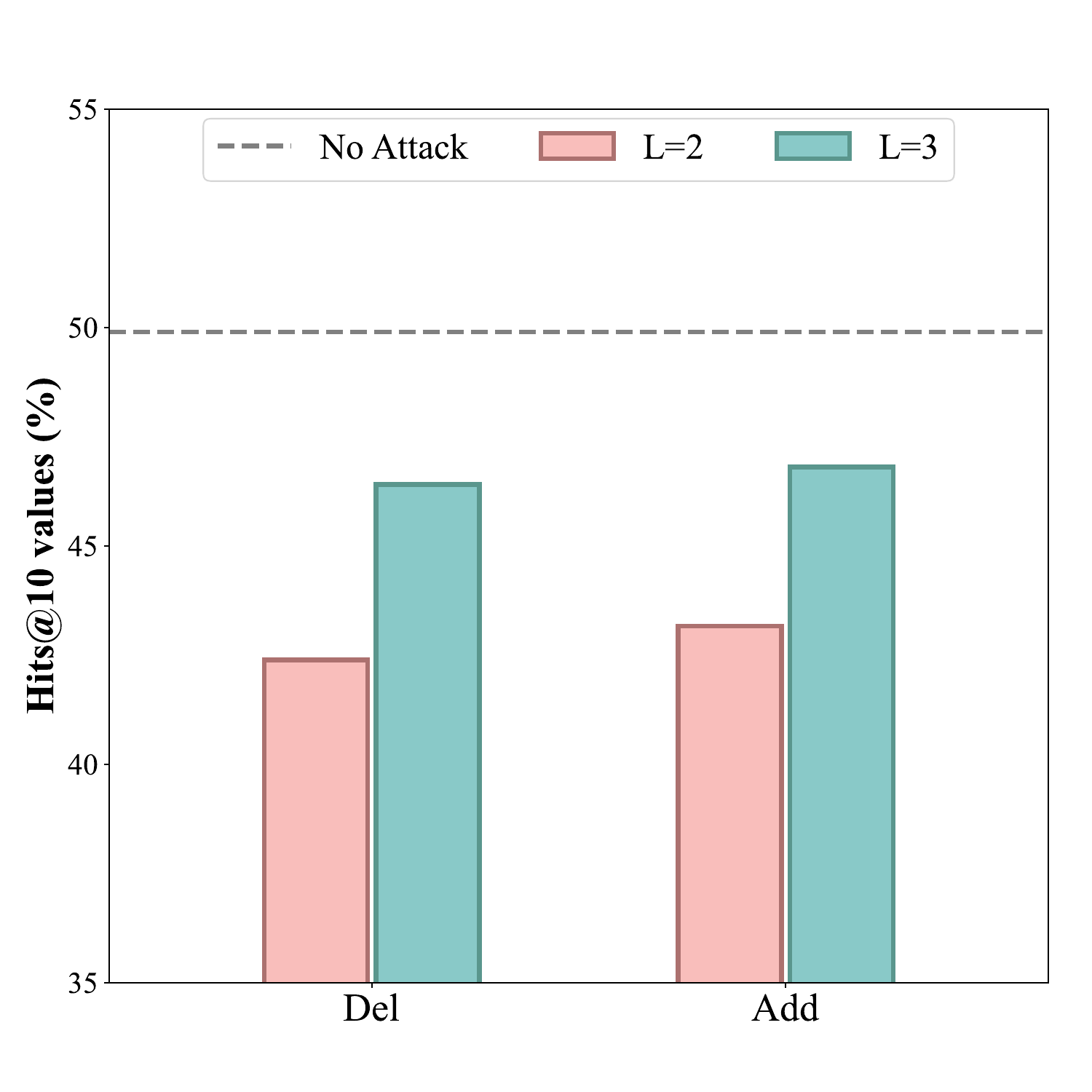}
    }
    \subfloat[DistMult]
    {
        \includegraphics[width=0.225\textwidth]{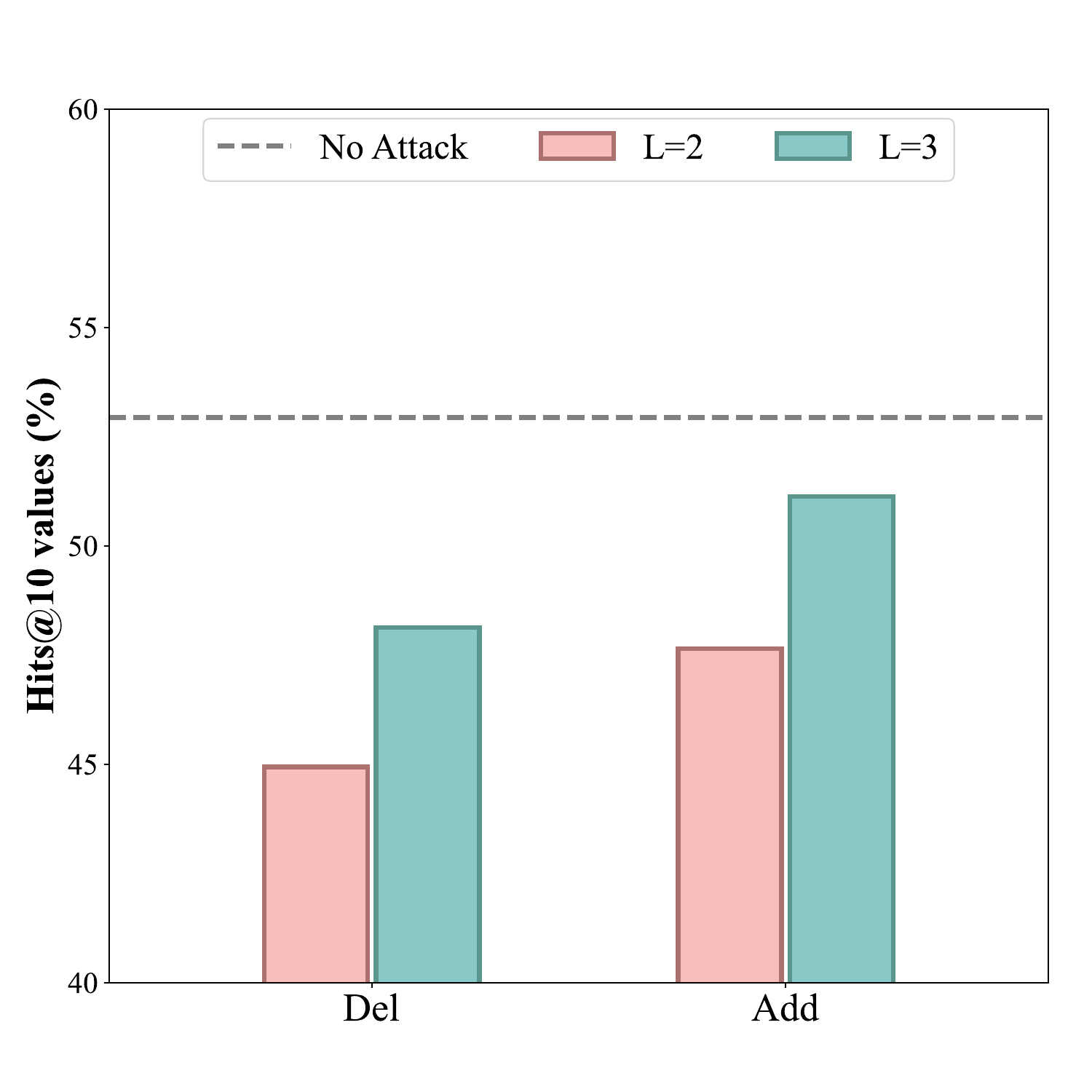}
    }
  \setlength{\abovecaptionskip}{0.1cm} 
  \caption{Results on WN18RR with rules of different lengths.}
  \label{fig:length}
  \Description{}
  \vspace{-0.4cm}
\end{figure}

\subsection{Ablation Study (\textbf{C3})}
\subsubsection{Employing Different Rule Extractors}
To estimate the influence of different rule extractors on attack efficiency, we also utilize rules learned by RNNLogic and compare the attack performances with the rules learned via NCRL. 
%Considering that RNNLogic compute rules' log likelihood score instead of confidence values, we treat the likelihood score as the same function to the confidence value. 
Similarly, logic rules for the deletion and addition attacks are selected based on the scores learned by RNNLogic.
The results are listed in Table \ref{tab:extractor}, where employing RNNLogic and NCRL show competitive performances on FB15k-237. However, on WN18RR, we find that adversarial attacks based on rules extracted from NCRL have superior performances. This reveals that NCRL could summarize more general patterns of WN18RR than RNNlogic, which indicates a potential application of our proposed attacks to distinguish the rule learning ability of KGE methods. 

\begin{table}[]
\setlength{\abovecaptionskip}{0.1cm}
\caption{Results of utilizing different predicate rewriting strategies for adversarial addition attacks. \textit{Corr} denotes the strategy based on the relation correlation values.}
\setlength{\abovecaptionskip}{0.1cm}
\resizebox{0.45\textwidth}{!}{
\begin{tabular}{cccclcc}
\toprule
\multirow{2}{*}{\textbf{Dataset}}            &  \multirow{2}{*}{\textbf{Rewriting}}                   & \multicolumn{2}{c}{\textbf{TransE}}      & & \multicolumn{2}{c}{\textbf{DistMult}}      \\
\cmidrule{3-4} \cmidrule{6-7} 
 &          & \textbf{MRR}    & \textbf{Hits@10}  &  & \textbf{MRR}     & \textbf{Hits@10}   \\
\midrule
\multirow{3}{*}{\textbf{FB15k-237}} & Random & 27.79        & 46.11    & & 24.50             & 38.93    \\
\cmidrule{2-7}                                  
                                    & Corr   & 24.77        & 42.25    & & 23.90             & 38.26   \\
                                    & $\Delta$ & \cellcolor{pink!50}-3.02      & \cellcolor{pink!50}-3.86    & & \cellcolor{pink!50}-0.60             & \cellcolor{pink!50}-0.67   \\
\midrule
\multirow{3}{*}{\textbf{WN18RR}}    & Random       & 19.09        & 48.40    & & 39.85             & 51.47     \\
\cmidrule{2-7}                                    
                                    & Corr   & 15.36       & 43.17    & & 24.43             & 47.65   \\
                                    & $\Delta$ & \cellcolor{pink!50}-3.73      & \cellcolor{pink!50}-8.70          & & \cellcolor{pink!50}-15.42             & \cellcolor{pink!50}-3.82   \\
\bottomrule
\end{tabular}
}
\label{tab:rewrite}
\vspace{-0.4cm}
\end{table}

\subsubsection{Impact on the Length of Extracted Rules}
We investigate the impact of the length of extracted rules on the attack performances. As shown in Figure \ref{fig:length}, although rules with length $L=3$ exhibit more complex patterns, it brings an inferior reduction of link predictions for both TransE and DistMult.
This implies that the conjunction of two atoms is the most common symbolic pattern in KGs.

\subsubsection{Effectiveness of Predicate Rewriting Strategy}
In the adversarial addition attack, we disrupt valid rules by rewriting one predicate of the rules' body based on the relation correlation values. To verify the effectiveness of this module, we compare it to the random replacement of relations. As shown in Table \ref{tab:rewrite}, disrupting rules based on the correlation values makes both TransE and DistMult get worse link prediction results. Especially on WN18RR, in the case of employing correlation values, DistMult shows a decrement of 15.42 in the MRR metric compared to the result obtained using random replacement. This well demonstrates the effectiveness of our predicate rewriting strategy. 

\section{Related Works}
\subsection{Knowledge Graph Embedding}
Simple KGE aims to derive the plausibility of a single fact and represents entities and relations in a low-dimensional space. One typical way to develop KGE models is treating the relation as the translation or mapping operation from the head entity to the tail, like TransE~\cite{bordes_2013} and RotatE~\cite{sun2018rotate}. Another tendency\cite{2014Embedding,pmlr-v48-trouillon16, dettmers2018convolutional} for matching latent semantics within the entire triple also shows great success. Regarding the topology of the whole KG, R-GCN~\cite{10.1007/978-3-319-93417-4_38} and CompGCN~\cite{Vashishth2020Composition-based} implement GNN to enhance the representations of entities and relations. Apart from only learning precise representations, many works seek to extract logic rules~\cite{10.5555/3294771.3294992, 10.1145/3308558.3313612, cheng2023neural} to infer missing links in a more convincing way. Despite the effectiveness of KGE, KGE models are fragile to unreliable data or disturbance in KGs and eventually learn biased embeddings \cite{9458733, 10.1145/3543507.3583203}.

In this paper, different from existing works about data attacks on KGs which only test simple KGE models, we take account of a wide range of KGE models including GNN-incorporated models and rule-engaged models to make a comprehensive evaluation.

\subsection{Adversarial Attacks against KGE}
Adversarial attacks against KGE usually conduct perturbations on the KG structure to bring a negative impact on link prediction. Specifically, existing works aim at proposing a data poisoning attack to reduce the performances of link prediction over selected target triples for simple KGE. CRIAGE~\cite{pezeshkpour-etal-2019-investigating} attempts to study the robustness and interoperability of KGE through adversarial addition and deletion according to the change of triple scores. With the help of Taylor expansion, CRIAGE can efficiently approximate these changes without retraining. Zhang et al.~\cite{ijcai2019p674} poison the KG in both direct and indirect ways to minimize the plausibility of targeted facts. Bhardwaj et al.~\cite{DBLP:conf/emnlp/BhardwajKCO21} propose to decide adversarial modifications via measuring the instance attribution, such as cosine similarity and $l_2$-distance between targeted triples and adversaries. Focusing on relational inference patterns including symmetry, inversion and composition, Bhardwaj et al.~\cite{DBLP:conf/acl/BhardwajKCO20} add adversarial triples which fit the relation pattern that can help get decoy triples. Once such adversarial additions have the high potential of inferring decoys, the ranks of the targets can be degraded in sequence. 

All the above attack strategies depend on detailed knowledge of KGE models. To get rid of the reliance on KGE, Betz et al.~\cite{betz2022adversarial} proposed to find explanations of the targets by logic rules. However, when adding noisy triples, they just adapt random replacements of entities in true triples but ignore the influence of logic rules, which may generate ridiculous perturbations. In general, although existing methods can well decrease the ranks of target triples, a more realistic way to develop a target-agonized attack to alleviate global performances overall test triples has not been fully studied.

\subsection{Adversarial Attacks on Graphs} Comparing with the limited discussion on the link prediction task in KGs, there has been much advancement in approaches~\cite{10.1145/3219819.3220078, 10.1145/3447548.3467416, 10.1145/3460120.3484796} that consider adversarial untargeted attacks on graphs. They violate edges in graphs or even perturb node features leading graph representation learning models like GNN to get misclassification results. However, it can hardly directly employ these methods in poisoning KGs due to several fundamental differences from our work. Firstly, the majority of them conduct perturbations on homogeneous graphs, while KGs are highly heterogeneous graphs whose edges illustrate various kinds of relations with abundant semantics. Secondly, they usually consider supervised node classification tasks or graph classification tasks for benchmarking attacks, while we evaluate link prediction tasks without explicit labels.

\section{Conclusion}
In this paper, we introduce a more practical scenario for adversarial attacks on KGEs in the untargeted attack setting, where all test triples keep unknown for the attack procedure. In regard to the association between logic rules and KGs, we develop attack strategies leveraging logic rules. The experimental results on two datasets over seven representative KGE methods verify the effectiveness of our proposed attacks. The comparison results demonstrate obvious superiority over baseline attackers. Meanwhile, we find that most KGE methods are more robust to addition attacks to deletion attacks, while Rule-based methods are sensitive to adversarial additions, where they probably capture negative rules. As for the future study, although corrupting the extracted logic rules has achieved good performances in damaging the general semantics of the KG, we will explore the influence of perturbing KG high-level semantics like entity hierarchies, relation hierarchies, and relation properties from ontological schemas.

%%
%% The acknowledgments section is defined using the "acks" environment
%% (and NOT an unnumbered section). This ensures the proper
%% identification of the section in the article metadata, and the
%% consistent spelling of the heading.
\begin{acks}
This work was supported by National Key Research and Development Program of China (2022YFC3303600), National Natural Science Foundation of China (62137002, 62293553, 62176207), “LENOVO-XJTU" Intelligent Industry Joint Laboratory Project, the EPSRC project Concur award (EP/V050869/1), Zhejiang Provincial Natural Science Foundation of China (LQ24F020034).

\end{acks}

%%
%% The next two lines define the bibliography style to be used, and
%% the bibliography file.
\bibliographystyle{ACM-Reference-Format}
\bibliography{sample-base}

%%
%% If your work has an appendix, this is the place to put it.
%%\appendix

%%\section{Research Methods}

%%\subsection{Part One}

\end{document}